\documentclass[a4paper,fleqn]{cas-dc}

\usepackage[numbers]{natbib}
\usepackage{amsmath,amssymb}
\usepackage{graphicx}
\usepackage{colortbl}
\usepackage{booktabs}
\usepackage{caption}
\usepackage{placeins}
\usepackage{float}
\def\tsc#1{\csdef{#1}{\textsc{\lowercase{#1}}\xspace}}
\tsc{WGM}
\tsc{QE}

\definecolor{rankone}{HTML}{005096}
\definecolor{ranktwo}{HTML}{3C8CBE}
\definecolor{rankthree}{HTML}{BED2E6}
\newcommand{\rankonecell}[1]{\cellcolor{rankone}\textcolor{white}{\textbf{#1}}}
\newcommand{\ranktwocell}[1]{\cellcolor{ranktwo}\textcolor{white}{\textbf{#1}}}
\newcommand{\rankthreecell}[1]{\cellcolor{rankthree}{#1}}

\begin{document}
\let\WriteBookmarks\relax
\def\textpagefraction{.001}

\shorttitle{UltraFast-LiNET}    

\shortauthors{Chen et al.}  

\title [mode = title]{UltraFast-LiNET: Light-weight Multi-Scale Shift Convolutional Network for Real-Time Low-Light Image Enhancement}  



%

\author[a]{Yuhan Chen}

\author[a]{Yicui Shi}

\author[a]{Guofa Li}

\cormark[1]

\ead{liguofa@cqu.edu.cn}
\orcidauthor{0000-0002-7889-4695}{Guofa Li}

\author[b]{Guangrui Bai}

\author[a]{Wenxuan Yu}

\author[a]{Ying Fang}

\author[c]{Wenbo Chu}

\author[d]{Keqiang Li}


\affiliation[a]{organization={College of Mechanical and Vehicle Engineering, Chongqing University},
            addressline={}, 
            city={Chongqing},
            postcode={400044}, 
            state={},
            country={China}}

\affiliation[b]{organization={School of Engineering Science, University of Science and Technology of China},
            addressline={}, 
            city={Hefei},
            postcode={230026}, 
            state={},
            country={China}}

\affiliation[c]{organization={National Innovation Center of Intelligent and Connected Vehicles},
            addressline={}, 
            city={Beijing},
            postcode={100089}, 
            state={},
            country={China}}

\affiliation[d]{organization={School of Vehicle and Mobility, Tsinghua University},
            addressline={}, 
            city={Beijing},
            postcode={100084}, 
            state={},
            country={China}}

\cortext[1]{Corresponding author}



\begin{abstract}
Addressing the urgent need for high-performance real-time low-light image enhancement on resource-constrained edge devices in low-illumination scenarios such as nighttime and tunnels, this paper presents UltraFast-LiNET, an ultra-lightweight network designed for extreme efficiency. The proposed network introduces Dynamic Shift Convolution, DSConv, a highly compact operation with only 12 learnable parameters. By using DSConv to construct a Multi-Scale Shift Residual Block, MSRB, UltraFast-LiNET effectively enlarges the receptive field while maintaining an exceptionally small parameter count. To alleviate gradient instability during the training of lightweight models, we further propose a multi-level gradient-aware loss function that strengthens supervision signals across different feature levels. UltraFast-LiNET supports flexible parameter configurations, enabling the model size to be reduced to as few as 36 learnable parameters. Remarkably, with only 180 learnable parameters, UltraFast-LiNET achieves a PSNR of 19.81~dB on the LOL dataset, surpassing the current state-of-the-art by 0.08~dB. Experimental results demonstrate that the proposed method achieves an effective balance between millisecond-level real-time processing and superior enhancement quality under extreme computational constraints. These results highlight its chip-level computational efficiency and broad prospects for edge-side deployment.
\end{abstract}


\begin{highlights}
\item An ultra-lightweight DSConv operator with only 12 learnable parameters is proposed.
\item A multi-scale shift residual block is designed for efficient receptive-field expansion.
\item A multi-level gradient-aware loss is introduced for robust real-time low-light enhancement.
\end{highlights}


\maketitle

\section{Introduction}
\label{sec:introduction}

\begin{figure*}[!t]
  \centering
  \includegraphics[width=\linewidth]{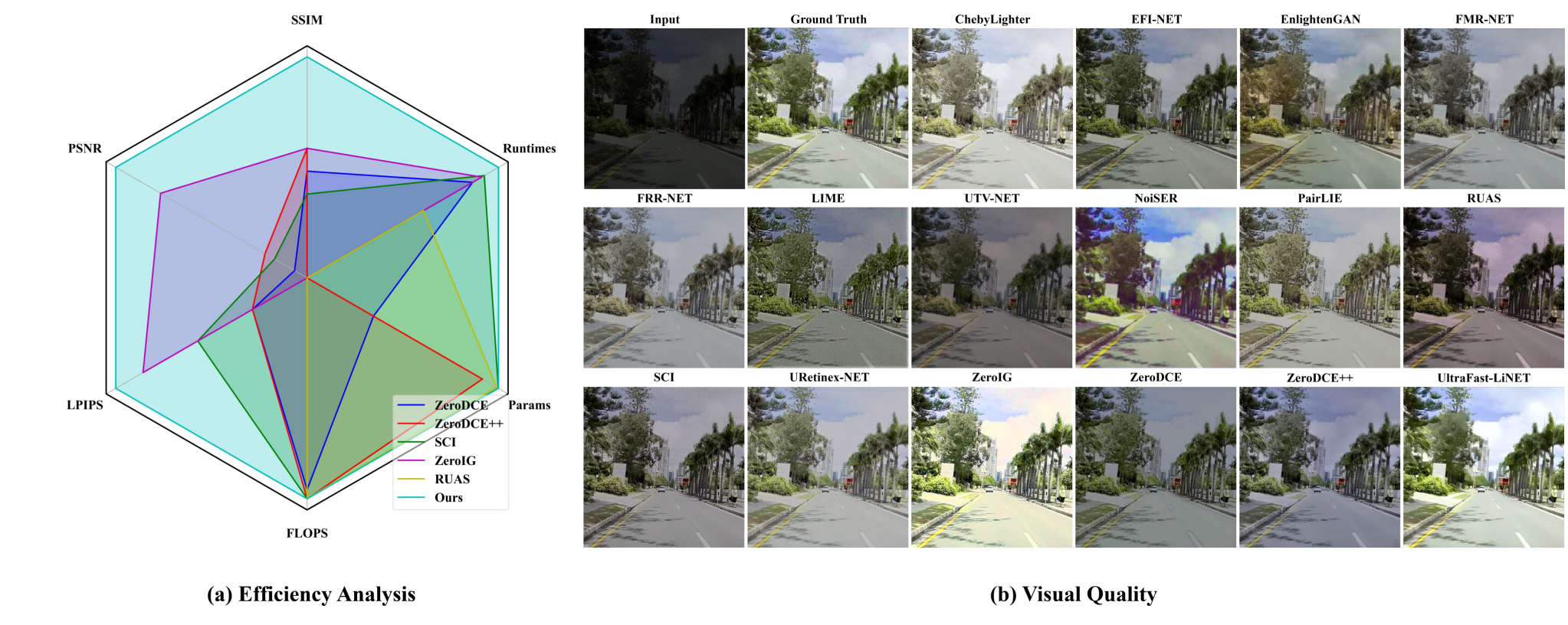}
  \caption{Comparison of performance and efficiency with existing state-of-the-art methods. (a) Image quality metrics (PSNR, SSIM, LPIPS) and computational efficiency (Params, FLOPs, Runtimes).}
  \label{fig1}
\end{figure*}

In practical outdoor visual applications, low-light images are typically defined as those captured by front-end vision sensors under poorly illuminated conditions, such as at night, in tunnels, or during dawn and dusk. Insufficient ambient lighting and inadequate exposure cause color distortion and loss of detail in image data, thereby impairing the performance of downstream vision tasks, e.g., detection, recognition, and tracking \citep{Islam2024LOLIStreet,Li2021Survey,Zhao2025LowLightVisionSurvey}.

Existing solutions primarily focus on two aspects: hardware and software algorithms. Hardware-based approaches are mainly targeted at high-end devices, where the introduction of costly sensors, including high dynamic range (HDR) sensors, near infrared (NIR) sensors, and event cameras, significantly enhances visual processing capabilities in low-light conditions \citep{Bai2025LiteIE,Guo2024HawkDrive,Liu2024NTIRELowLight}. However, due to the high cost of hardware and sensors, substantial power consumption, and limited computational capacity of embedded platforms, these solutions struggle to meet the efficiency and low-power-consumption requirements of large-scale real-world visual system applications \citep{Ozcan2023FLIGHT,Zhang2024BinarizedRawVideo}. Consequently, software-based solutions, primarily relying on low-light image enhancement (LLIE) algorithms, have become the optimal choice for most mobile and edge device applications.

For the low-light image enhancement problem, early algorithmic studies were primarily devoted to traditional image signal processing (ISP) methods, such as global or local histogram equalization and Retinex-based enhancement algorithms \citep{Pisano1998CLAHE,Abdullah2007DHE,Pizer1987AHE,Land1971Retinex,Guo2016LIME}. Although issues of brightness and contrast are partially alleviated, reliance on manually designed feature extractors and parameter configurations hinders adaptive balancing between detail reconstruction and noise suppression in dynamic, complex real-world scenarios, thereby limiting generalization capability \citep{Li2021ZeroDCE}. In contrast, deep learning-based enhancement methods have, in recent years, been shown to provide superior scene adaptability and the potential for overall performance improvement through autonomous learning of robust feature representations and nonlinear mappings from large-scale data \citep{Guo2020ZeroDCE,Ma2022SCI,Liu2021RetinexUnrolling,Jiang2021EnlightenGAN,Chen2024FMR,Chen2024FRR,Zheng2021AdaptiveUnfolding,Pan2022ChebyLighter,Liu2022EFINet,Fu2023SimpleEnhancer,PerezZarate2023LoLiIEA,Zhang2024NoiseSelfRegression,Wu2022URetinex,Shi2024ZEROIG}.

In the LLIE field, deep learning methods primarily follow three paradigms: supervised learning, unsupervised or zero-reference learning, and generative enhancement. In supervised learning, illumination mappings are learned from strictly paired low-light and normal-light images, and early network architectures are predominantly based on convolutional neural networks (CNNs) \citep{Lore2017LLNet,Wei2018RetinexNet}. With the evolution of network architecture design, new architectures based on Transformer and Mamba are gradually replacing CNNs as the mainstream solution for supervised learning \citep{Jiang2023StageTransformer,Wen2025DualAttentionTransformer,Bai2024RetinexMamba,Chen2025RetinexRawMamba,SeeingBeyondDarkness2026}. The core limitation of these methods lies in their strong dependence on paired data, which restricts their generalization ability and scene adaptability. By contrast, unsupervised and semi-supervised methods free models from the constraints of paired data. Representative approaches, with GANs and diffusion models as typical examples, enhance generalization and scene adaptability through generative learning mechanisms \citep{Li2021ZeroDCE,Guo2020ZeroDCE,Jiang2021EnlightenGAN,Ghosh2019IEGAN,Du2025UIEDP,Wang2025LLDiffusion,Xu2025LearnableFeature,Guo2024OneRestore}. However, deployment efficiency on edge devices is hindered by their large parameter scales, which in turn restricts application flexibility.

Although deep learning has significantly advanced the field of low-light image enhancement (LLIE), current state-of-the-art (SOTA) methods face challenges in achieving an effective balance between computational efficiency and enhancement performance due to high model complexity and deployment constraints. Consequently, efficient real-time inference on resource-constrained edge platforms remains difficult to achieve. To address the aforementioned challenges, this paper proposes UltraFast-LiNET, an ultra-lightweight network designed for real-time low-light image enhancement, aimed at building a dedicated lightweight backbone architecture for resource-constrained embedded platforms. The network significantly optimizes memory usage and computational efficiency for downstream tasks. Its performance comparison against SOTA methods is quantitatively illustrated in Fig.~\ref{fig1}. The core innovations of the network are summarized as follows:

\begin{enumerate}
\item A dynamic shift convolution kernel (DSConv) containing merely 12 learnable parameters is proposed in UltraFast-LiNET. Through integration of DSConvs with varying shift distances, a plug-and-play multi-scale shift residual block (MSRB) is constructed. This module achieves significantly expanded receptive fields with extremely low parameter overhead. Using MSRB as the fundamental building block, an ultra-lightweight backbone network is developed specifically for real-time applications.
\item To enhance the overall learning performance of UltraFast-LiNET, a novel multi-level gradient-aware loss function is introduced to effectively supervise gradient variations of deep features at multiple levels.
\item An ultra-lightweight backbone network is constructed with MSRB as the core component, reducing the parameter count by more than 40\% compared with state-of-the-art methods. UltraFast-LiNET simultaneously achieves performance comparable to state-of-the-art approaches on multiple datasets.
\end{enumerate}

\section{Related Work}
\label{sec:related_work}

With the growing prevalence of edge computing and mobile applications, lightweight low-light image enhancement algorithms for resource-constrained embedded hardware devices have attracted increasing research attention. The lightweight design in this domain primarily focuses on two aspects: the development of lightweight algorithms and the structural optimization of fundamental modules, such as convolution kernels.

\subsection{Algorithm lightweight design}
\label{subsec:algorithm_lightweight_design}

In the field of low-light image enhancement, supervised methods are typically built upon backbone networks such as CNN, Transformer, or Mamba, where paired data are directly utilized for supervised learning \citep{Lore2017LLNet,Wei2018RetinexNet,Jiang2023StageTransformer,Wen2025DualAttentionTransformer,Bai2024RetinexMamba,Chen2025RetinexRawMamba,Hu2025LightenMST,Zhang2022DCC,Xu2022SNRAware}. In contrast, unsupervised methods are mainly developed on architectures such as deconvolutional networks or U-Net, in which illumination transformations are simulated through data generation strategies \citep{Li2021ZeroDCE,Guo2020ZeroDCE,Jiang2021EnlightenGAN,Ghosh2019IEGAN,Du2025UIEDP,Wang2025LLDiffusion}. Due to the general-purpose nature of their frameworks, these models usually involve a large number of parameters. In addition, performance improvement often requires the introduction of new branches or structures, which further challenges deployment on edge-side devices. As a result, lightweight algorithm design has gradually emerged as a central direction. In basic lightweight design, minimal parameter size and computational cost are considered, making it suitable for resource-constrained edge devices. For instance, a lightweight Transformer model with only 90K parameters is proposed by Cui et al.~\citep{Cui2022LightweightTransformer}. Weng et al. combine dark/bright channel priors and gamma correction to achieve low-light enhancement using deep learning, requiring only 25K parameters \citep{Weng2024ChannelPrior}. Guo et al. frame low-light image enhancement as a task of image-specific curve estimation with deep networks, and proposed ZeroDCE, which enables fast and effective image enhancement with only 10K parameters \citep{Li2021ZeroDCE,Guo2020ZeroDCE}. Ma et al. design a fast, flexible, and robust low-light enhancement framework, SCI, by integrating a cascaded illumination learning process with weight sharing, requiring only 300 learnable parameters \citep{Ma2022SCI}.

Although basic lightweight design enables significant reduction in parameter count, its effectiveness is often limited in complex scenarios, where challenges such as color distortion, high noise, and detail loss persist. Consequently, lightweight algorithms tailored to specific low-light enhancement scenarios have emerged. For instance, Perez-Zarate et al. propose LoLi-IEA, a two-stage method that first classifies luminance levels and then performs enhancement based on the classification results, focusing on detail extraction and making it suitable for real-time applications \citep{PerezZarate2023LoLiIEA}. Brateanu et al. optimize color distortion and detail loss with a transformer network based on the YUV color space, achieving superior performance over the Retinex method, and making it suitable for real-time computer vision tasks \citep{Brateanu2025LYTNet}.

The performance of lightweight low-light image enhancement algorithms has been further improved by the integration of recent deep learning techniques, including Transformer, Mamba, and KAN \citep{Zhang2025BSMamba,KAN2025SeeInDark}. For instance, Wave-Mamba is proposed by Zou et al., which is based on wavelet domain and state-space models, and enables effective detail recovery as well as accurate color correction \citep{Zou2024WaveMamba}. Lu et al. guided by semantic priors, combine the KAN module to enhance high and low-frequency image signals, achieving noise suppression while enhancing the image \citep{Lu2025KANformer}. Zhang et al. propose STAR, a lightweight Transformer network, suitable for real-time image enhancement, which effectively improves image illumination and corrects white balance \citep{Zhang2021STAR}. Although these integrated technologies have made great progress in improving the performance of lightweight algorithms, most such designs still struggle to balance computational efficiency and inference performance, which limits their real-time applicability on edge devices \citep{Xue2025CLIPFourier}.

\subsection{Basic module structure optimization}
\label{subsec:basic_module_structure_optimization}

Structural optimization of fundamental modules typically prioritizes network pruning and the development of lightweight feature extraction modules, whereas more effective approaches have focused on reconstructing fundamental elements such as convolutional kernels. A primary objective in lightweight architectures is the reduction of kernel parameters while maintaining the receptive field and feature extraction capability. The fundamental paradigm of convolutional kernel lightweighting is primarily achieved through decomposition or grouped convolutions. For example, Xie et al. propose grouped convolutions, which reduce computational overhead by dividing the input channels into multiple groups and performing convolutions independently within each group, thus minimizing cross-channel interactions \citep{Xie2017ResNeXt}. Chollet et al. propose the well-known depthwise separable convolution, which decomposes the standard convolution into depthwise convolution, applied independently to each channel, and pointwise convolution, i.e., $1 \times 1$ convolution that fuses channels, significantly reducing computational complexity \citep{Chollet2017Xception}.

Fundamental paradigms are currently widely employed in deep learning architectures. However, they are predominantly concerned with extensive parameter minimization, frequently resulting in compromised accuracy or information redundancy. Consequently, optimization strategies for convolutional kernels aimed at these limitations are being increasingly developed. For instance, Zhang et al. introduce channel shuffle after grouped convolutions to address the problem of information isolation caused by channel partitioning, thereby enhancing feature interaction while maintaining low computational cost \citep{Zhang2018ShuffleNet}. Sandler et al. design a structure that first expands channels, then applies depthwise convolution, and finally compresses them, which avoids information loss in low-dimensional space and effectively improves accuracy \citep{Sandler2018MobileNetV2}. Han et al. generate ``Ghost'' feature maps through inexpensive operations to reduce redundant computations, addressing the issue of feature redundancy in depthwise separable convolutions \citep{Han2020GhostNet}. Huang et al. optimize the grouped structure by learning grouped convolutions to eliminate redundant connections, thereby improving efficiency and mitigating accuracy degradation \citep{Huang2018CondenseNet}. The algorithmic performance of lightweight convolutional kernels has been enhanced through the recent introduction of diverse dynamic and optimization mechanisms. These approaches effectively mitigate poor adaptability and unstable optimization in complex scenarios, resulting in substantial gains in both efficiency and robustness. For example, convolutional kernel weights are dynamically adjusted according to the input by Chen et al., addressing the adaptability limitation of static kernels while maintaining lightweight characteristics \citep{Chen2020DynamicConv}. Chan et al. employ low-rank decomposition to approximate convolutional kernels, which reduced parameters, mitigated high-dimensional redundancy, and improved computational speed \citep{Chan2015PCANet}. These advances highlight the effectiveness of lightweight kernel design, although a balance between module performance and computational cost is still required.

\section{Proposed Method}
\label{sec:proposed_method}

\begin{figure*}[!t]
  \centering
  \includegraphics[width=\textwidth]{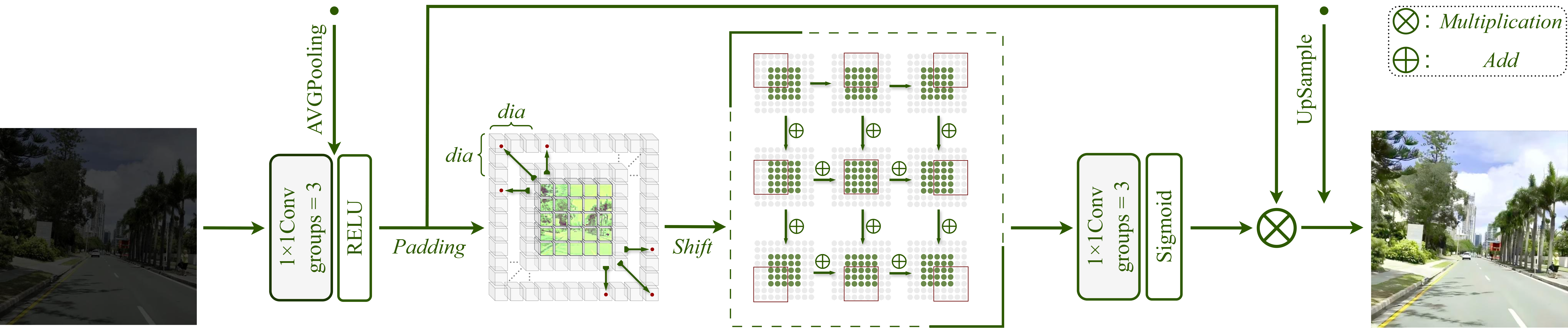}
  \caption{Lightweight architecture of DSConv, including spatial shifting and feature aggregation. Gated modulation is introduced in the output process to enable selective retention and enhancement of aggregated features.}
  \label{fig2}
\end{figure*}

\begin{figure*}[!t]
  \centering
  \includegraphics[width=\textwidth]{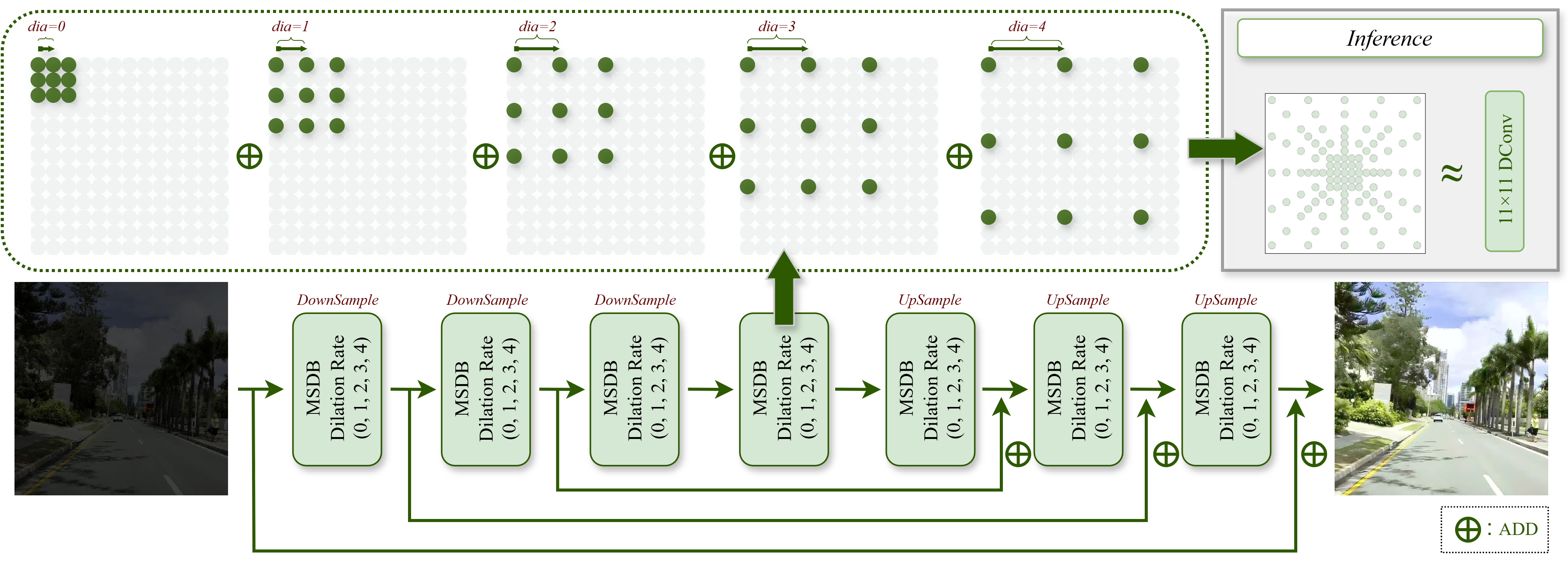}
  \caption{Overall architecture of UltraFast-LiNET. Progressive downsampling and upsampling are performed through multiple MSRBs, and dense skip connections are employed to preserve detailed information.}
  \label{fig3}
\end{figure*}

As illustrated in Fig.~\ref{fig2}, a dynamic shifting convolutional kernel (DSConv) with only 12 learnable parameters is presented in this section. Building upon this foundation, a multi-scale shifting residual block (MSRB) is designed and an ultra-lightweight network termed UltraFast-LiNET is constructed for real-time low-light image enhancement, as illustrated in Fig.~\ref{fig3}. Finally, the various loss functions employed are described, with particular emphasis placed on the newly proposed multi-level gradient-aware loss function.

\subsection{Dynamically shifted convolution kernel}
\label{subsec:dynamically_shifted_convolution_kernel}

To efficiently enlarge the receptive field and capture long-range dependencies without a significant increase in parameters or computational cost, a novel dynamic shifting convolutional kernel (DSConv) is introduced. As illustrated in Fig.~\ref{fig2}, DSConv expands the receptive field through parallel spatial shifting and feature aggregation, resembling the behavior of dilated convolution while maintaining distinct parameter efficiency and computational structure.

The core idea of DSConv is to replace explicit convolution weight multiplication with an effective combination of spatial shifting and lightweight channel transformations. Drawing inspiration from lightweight algorithms such as FasterNet and GhostNet, DSConv effectively leverages image feature redundancy to further optimize computational efficiency. This method fixes the number of feature channels at three for computation and realizes cross-channel feature fusion by superposing the new feature maps generated through shifting operations \citep{Wu2018Shift,Yu2015DilatedConv,Qin2024MobileNetV4,Chen2023FasterNet}.

Let the input tensor be denoted as $X \in \mathbb{R}^{B \times C \times H \times W}$, where $B$ is the batch size, $C$ is fixed at three channels, and the height and width of the input features are denoted by $H$ and $W$, respectively. The input tensor $X$ is first processed by a pointwise group convolution with a kernel size of $1 \times 1$ to produce two independent features for subsequent use, namely the offset feature $X_o$ and the residual feature $X_r$. Information fusion is performed only within each channel. The mathematical formulation is expressed as:
\begin{equation}
X_o = X_r = \mathrm{ReLU}\left(\mathrm{Conv}^{(1)}_{1 \times 1}(X)\right),
\label{eq:dsconv_initial}
\end{equation}
where $\mathrm{Conv}^{(1)}_{1 \times 1}$ denotes the first point-wise grouped convolutional layer with kernel size $1 \times 1$. $\mathrm{ReLU}$ represents the rectified linear unit activation function. Through this step, $X_o, X_r \in \mathbb{R}^{B \times C \times H \times W}$ are obtained.

To effectively enlarge the receptive field, the offset feature $X_o$ is used as the basis, and a series of parallel and regular spatial shifting operations are performed to simulate the coverage of convolutional kernels in the spatial dimension. Let the dilation rate of DSConv be denoted as $\mathrm{dia}$. The feature $X_o$ is first zero-padded, with $\mathrm{dia}$ pixels added along the four boundaries, i.e., top, bottom, left, and right, resulting in an expanded feature map $X_p$:
\begin{equation}
X_p = \mathrm{Pad}(X_o,[\mathrm{dia},\mathrm{dia},\mathrm{dia},\mathrm{dia}]).
\label{eq:padding}
\end{equation}
At this stage, $X_p \in \mathbb{R}^{B \times C \times (H+2\mathrm{dia}) \times (W+2\mathrm{dia})}$.
A $3 \times 3$ grid sampling is then performed on $X_p$ to generate nine shifted feature maps. Each feature map $X_{i,j}$ is obtained by cropping $X_p$ with a specific offset, where $i,j \in \{-1,0,1\}$ denote the offset indices along the vertical and horizontal directions. For mathematical consistency, the offset is defined as $(i \cdot \mathrm{dia}, j \cdot \mathrm{dia})$. At this stage, $X_{i,j}$ is expressed as:
\begin{equation}
\begin{split}
X_{i,j} &= X_p[:, :, (i+1)\mathrm{dia} : (i+1)\mathrm{dia}+H,\\
&\qquad\quad (j+1)\mathrm{dia} : (j+1)\mathrm{dia}+W].
\end{split}
\label{eq:shifted_feature}
\end{equation}

The nine shifted feature maps are summed element-wise and aggregated into a single feature map $X_{\mathrm{agg}}$. This process is expressed as:
\begin{equation}
X_{\mathrm{agg}}
=
\sum_{i \in \{-1,0,1\}}
\sum_{j \in \{-1,0,1\}}
X_{i,j}.
\label{eq:aggregation}
\end{equation}
This aggregation process effectively consolidates information from different spatial locations separated by the dilation rate $\mathrm{dia}$. Its effect is equivalent to that of a dilated convolution kernel with a kernel size of $3 \times 3$ and a dilation rate of $\mathrm{dia}$.

Finally, to enable the module to adaptively fuse the aggregated information, a gating mechanism is introduced. The aggregated feature $X_{\mathrm{agg}}$ is first processed by a second pointwise grouped convolution layer with a kernel size of $1 \times 1$, followed by a sigmoid activation function, generating an attention gating map $G \in \mathbb{R}^{B \times C \times H \times W}$. This process is expressed as:
\begin{equation}
G = \mathrm{Sigmoid}\left(\mathrm{Conv}^{(2)}_{1 \times 1}(X_{\mathrm{agg}})\right).
\label{eq:gating}
\end{equation}
The value range of the gating map $G$ is $(0,1)$, where a modulation weight is learned for each channel at every spatial location. Finally, the gating map $G$ is multiplied element-wise with the residual feature $X_r$ generated in the first step to obtain the final output $Y$, which is expressed as:
\begin{equation}
Y = G \otimes X_r.
\label{eq:dsconv_output}
\end{equation}
This gating mechanism allows the module to dynamically adjust the pathway of the original feature $X_r$ according to the importance of the aggregated information $X_{\mathrm{agg}}$, where selective retention and enhancement of information are achieved.

Since DSConv is often applied with upsampling or downsampling, average pooling and interpolation-based upsampling are incorporated into its architecture to facilitate network inference.

\begin{figure*}[t]
  \centering
  \includegraphics[width=\linewidth]{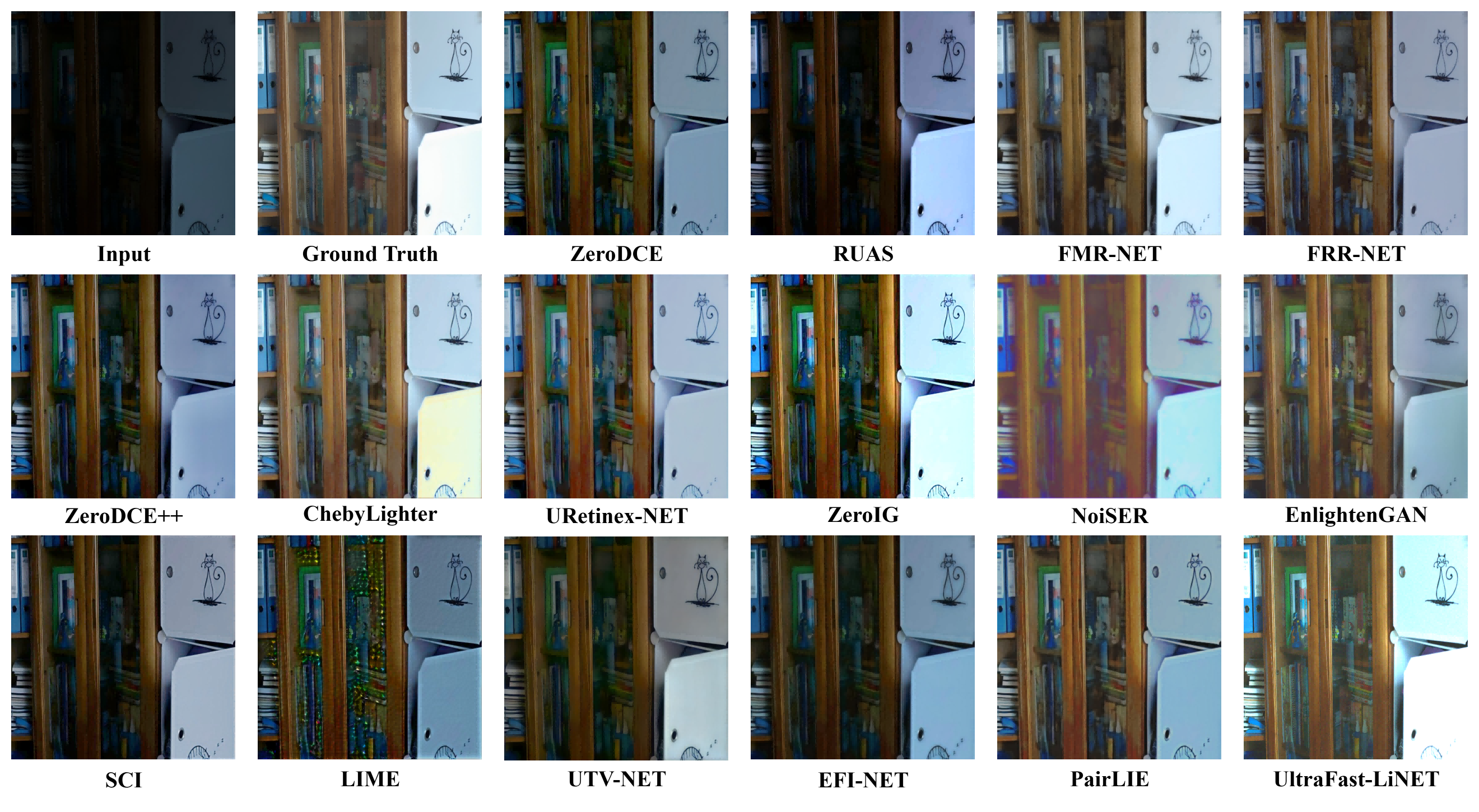}
  \caption{Visual comparisons of UltraFast-LiNET with state-of-the-art methods on the LOL benchmark datasets}
  \label{fig4}
\end{figure*}

\subsection{Dynamic shift convolution kernel as the basic operator}
\label{subsec:dynamic_shift_convolution_kernel_as_basic_operator}

To demonstrate the efficiency gain of the proposed DSConv, a comparison is conducted with the most functionally similar baseline model, dilated convolution. For parameter comparison, a dilated convolution with a kernel size of $3 \times 3$ and a dilation rate of $\mathrm{dia}$ is defined as a standard convolution layer to simulate the receptive field expansion achieved by DSConv. As described, the parameter count of DSConv is determined entirely by the weight and bias terms of its two $1 \times 1$ pointwise grouped convolutions. In UltraFast-LiNET, the convolutional weights in DSConv are initialized and scaled using the Kaiming Normal method. For a single $1 \times 1$ convolutional layer, the weight matrix $W \in \mathbb{R}^{C \times 1 \times 1 \times 1}$ and bias vector $b \in \mathbb{R}^{C}$ are initialized as follows:
\begin{equation}
\begin{aligned}
W_c \sim \mathcal{N}(0,\sigma^2) \cdot s,\quad
b_c = 0,\quad
\mathrm{for}\ c = 1,2,\ldots,C,
\end{aligned}
\label{eq:kaiming_initialization}
\end{equation}
where $\sigma$ is determined by Kaiming Normal, and $s=0.1$ denotes the scaling factor. Therefore, the sizes of the weight parameter $P_W$ and the bias parameter $P_b$ in a single convolution layer of DSConv are calculated as follows:
\begin{align}
P_W &= C_{\mathrm{out}} \times \frac{C_{\mathrm{in}}}{\mathrm{groups}} \times K_H \times K_W \nonumber\\
    &= C \times \frac{C}{C} \times 1 \times 1 = C,
\label{eq:weight_parameter}
\end{align}
\begin{equation}
P_b = C_{\mathrm{out}} = C,
\label{eq:bias_parameter}
\end{equation}
where $K$ is the convolution kernel size. Therefore, we can quickly deduce that the total number of parameters for a single DSConv is:
\begin{equation}
P_{\mathrm{DSConv}}
=
2 \times (P_W + P_b)
=
4C.
\label{eq:dsconv_parameter}
\end{equation}
The total number of parameters of the dilated convolution $P_{\mathrm{DConv}}$ is expressed as:
\begin{align}
P_{\mathrm{DConv}} &= P_W + P_b \nonumber\\
&= C_{\mathrm{out}} \times C_{\mathrm{in}} \times K_H \times K_W + C_{\mathrm{out}} \nonumber\\
&= C \times C \times 3 \times 3 + C = 9C^2 + C.
\label{eq:dilated_parameter}
\end{align}

\begin{figure*}[!t]
  \centering
  \includegraphics[width=\linewidth]{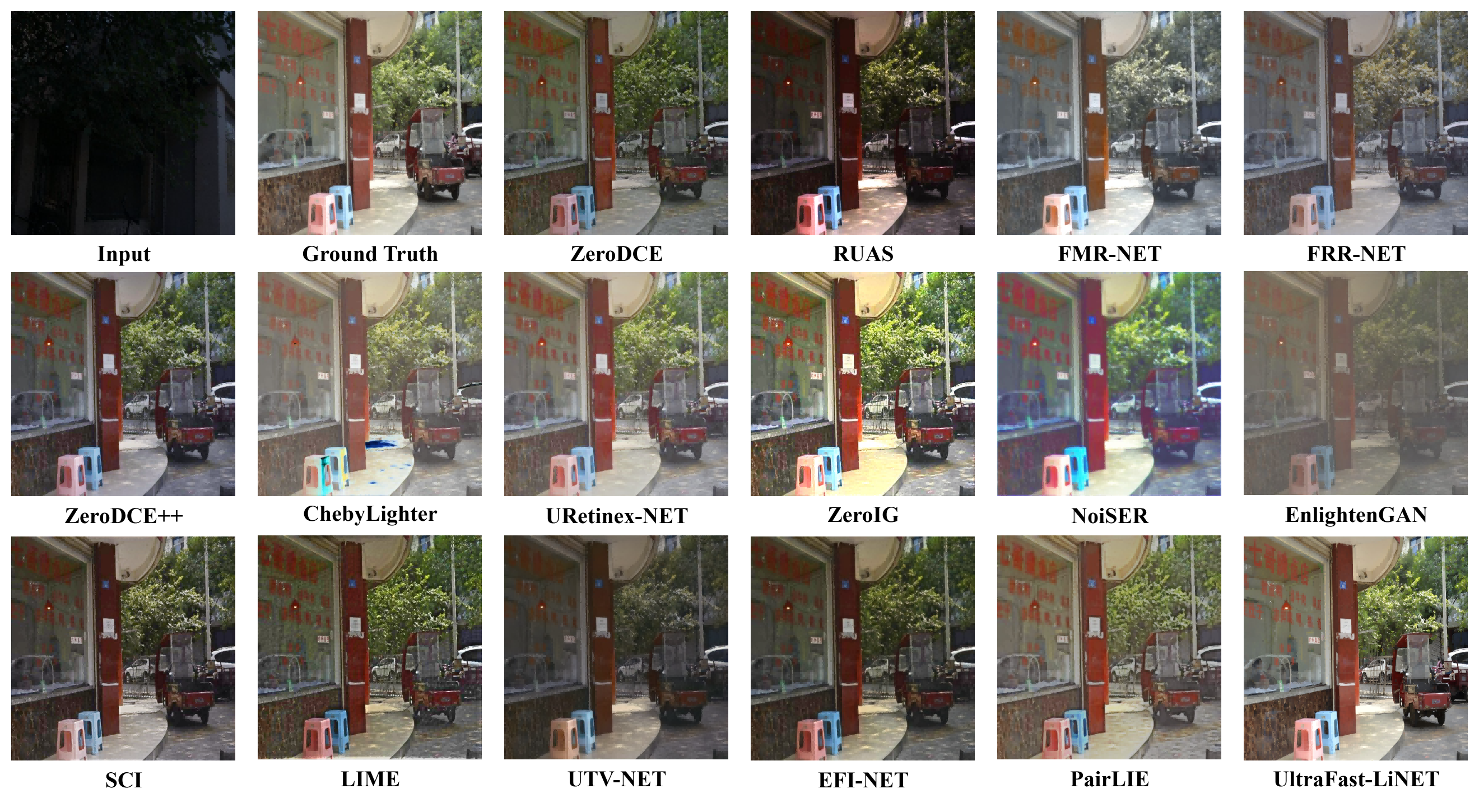}
  \caption{Visual comparisons of UltraFast-LiNET with state-of-the-art methods on the LSRW-HUAWEI benchmark datasets}
  \label{fig5}
\end{figure*}

\begin{figure*}[!t]
  \centering
  \includegraphics[width=\linewidth]{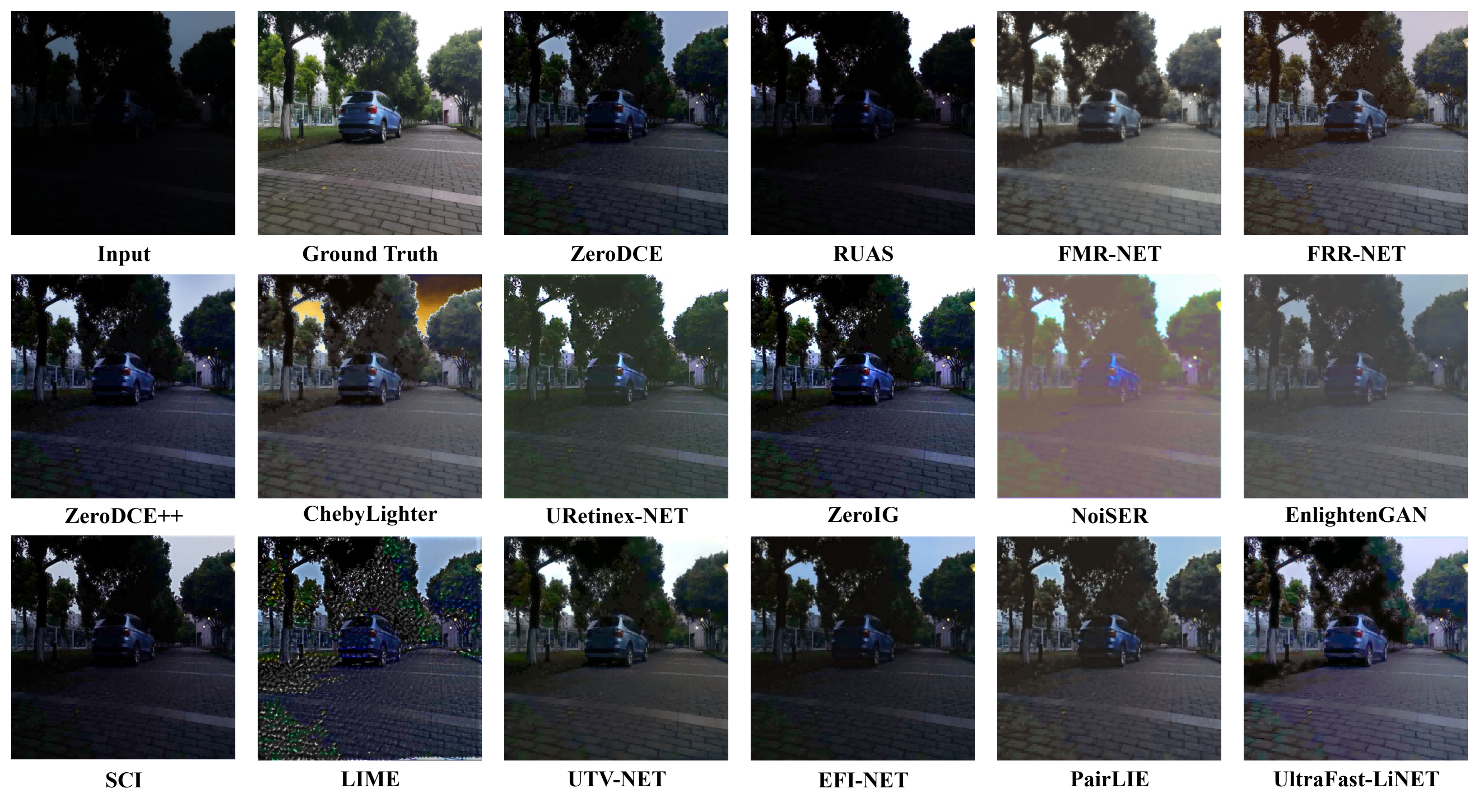}
  \caption{Visual comparisons of UltraFast-LiNET with state-of-the-art methods on the LSRW-NIKON benchmark datasets}
  \label{fig6}
\end{figure*}

As shown above, the number of parameters of dilated convolution is much higher than that of DSConv. When $C=256$, the number of parameters of DSConv is about $1/576$ of that of dilated convolution.

In terms of computational complexity, DSConv primarily consists of two $1 \times 1$ pointwise grouped convolutions, one addition aggregation, and one gated multiplication. The total computational complexity, excluding the activation function, can be expressed as:
\begin{equation}
\resizebox{0.96\columnwidth}{!}{$\displaystyle
\begin{aligned}
F_{\mathrm{DSConv}} &=
\underbrace{(4CHW)}_{2\times\mathrm{Conv}_{1\times1}^{(1)}}
+
\underbrace{(8CHW)}_{\mathrm{Aggregation}}
+
\underbrace{(2CHW)}_{\mathrm{Conv}_{1\times1}^{(2)}}
+
\underbrace{(CHW)}_{\mathrm{Mul}}
\\
&= 15CHW.
\end{aligned}
$}
\label{eq:dsconv_flops}
\end{equation}
In contrast, the computational complexity of dilated convolution mainly comes from convolution operation and bias addition. Its total computational complexity $F_{\mathrm{DConv}}$ can be expressed as:
\begin{equation}
F_{\mathrm{DConv}}
=
18C^2HW + CHW.
\label{eq:dilated_flops}
\end{equation}
From the above analysis, the computational cost of dilated convolution is also much higher than that of DSConv. When $C=256$, the computational cost of DSConv is approximately $1/307$ of that of dilated convolution, where $C$ denotes the number of channels.

\subsection{Network structure}
\label{subsec:network_structure}

Fig.~\ref{fig3} illustrates the deep auto-encoder architecture of UltraFast-LiNET, which is centered on progressive downsampling and upsampling via multi-scale feature blocks. Dense skip connections are incorporated for detail retention. The network is composed predominantly of plug-and-play multi-scale shifted residual blocks (MSRBs).

The MSRB is the core feature processing module in our design. Its central idea is to integrate multiple DSConvs with different dilation rates in parallel to efficiently capture multi-scale contextual information without introducing additional parameter overhead. A single MSRB consists of $\kappa$ parallel submodules $\{\mathrm{DSConv}_{\mathrm{dia}}\}_{\mathrm{dia}=0}^{\kappa-1}$, where $\kappa=5$ in this work. Given the input feature $X \in \mathbb{R}^{B \times C \times H \times W}$, the processing of this feature by the module is formulated as:
\begin{align}
\mathrm{MSRB}(X) &= \sum_{\mathrm{dia}=0}^{\kappa-1} g_{\mathrm{dia}}(X) \nonumber\\
&= \sum_{\mathrm{dia}=0}^{\kappa-1} \mathrm{DSConv}_{\mathrm{dia}}(X).
\label{eq:msrb}
\end{align}

\begin{figure*}[!t]
  \centering
  \includegraphics[width=\linewidth]{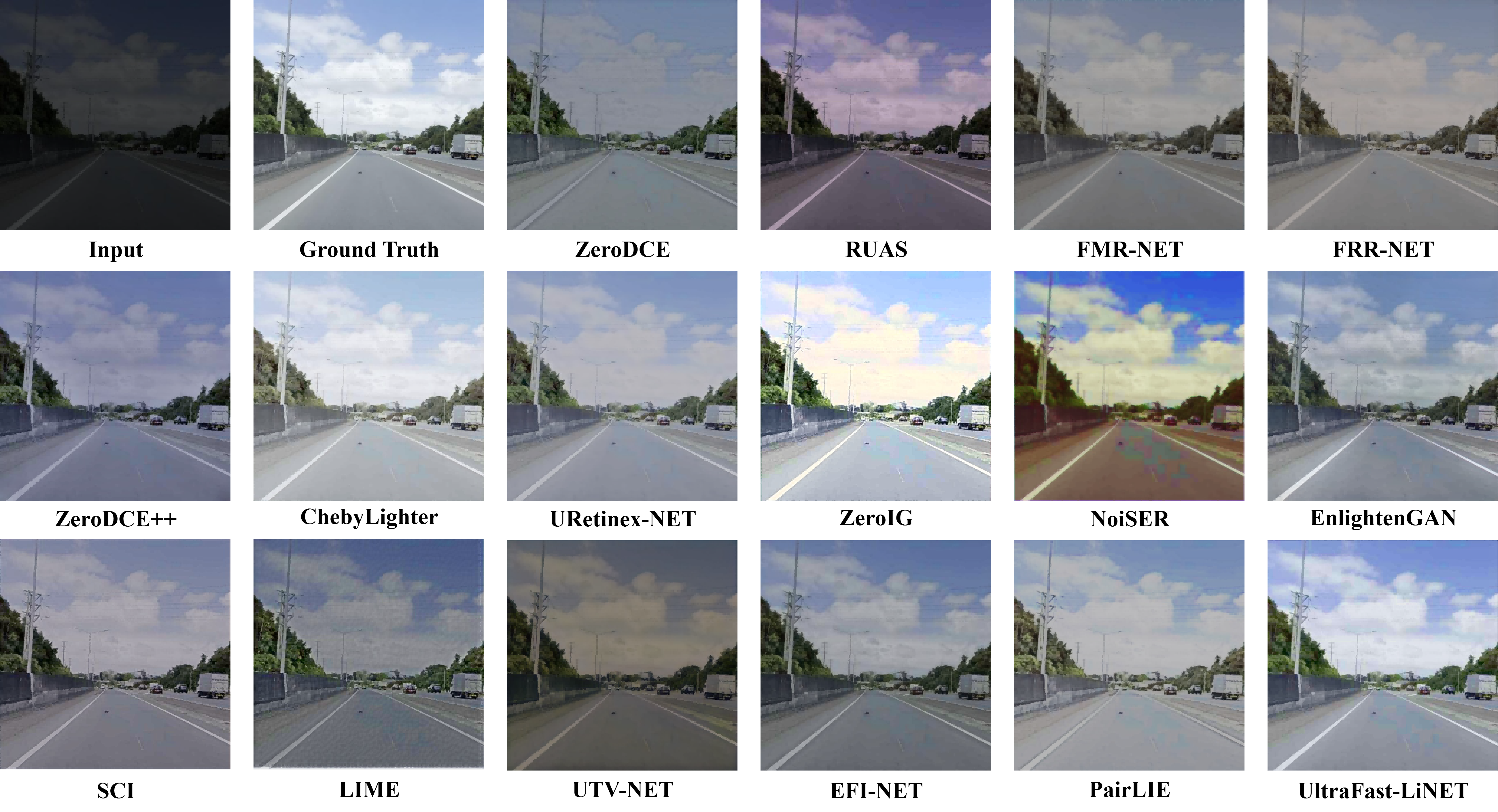}
  \caption{Visual comparisons of UltraFast-LiNET with state-of-the-art methods on the LOLI-Street benchmark datasets}
  \label{fig7}
\end{figure*}

This process indicates that the output of the module is the sum of the outputs from all parallel subpaths. This design enables the network to perceive multiple patterns ranging from fine local features with $\mathrm{dia}=0$ to broader contextual information with $\mathrm{dia}=4$, and its receptive field is equivalent to that of an $11 \times 11$ dilated convolution. Based on this scheme, downsampling and upsampling functions are incorporated into DSConv, from which two variants of the MSRB are derived, namely $M^{\downarrow}(X)=\sum_{\mathrm{dia}=0}^{\kappa-1}g_{\mathrm{dia}}^{\downarrow}(X)$ and $M^{\uparrow}(X)=\sum_{\mathrm{dia}=0}^{\kappa-1}g_{\mathrm{dia}}^{\uparrow}(X)$.
In the upsampling stage, $M^{\uparrow}(X)$ is responsible for upsampling and multi-scale feature fusion, and it also performs the integration of information from skip connections. It receives two inputs: the feature $X_{\mathrm{dec}}$ from the previous decoding layer and the skip connection feature $X_{\mathrm{enc}}$ from the encoder. The upsampling module is further expressed as:
\begin{equation}
M^{\uparrow}(X_{\mathrm{dec}},X_{\mathrm{enc}})
=
\sum_{\mathrm{dia}=0}^{\kappa-1}
\left(
g_{\mathrm{dia}}^{\uparrow}(X_{\mathrm{dec}})
\oplus
X_{\mathrm{enc}}
\right).
\label{eq:upsampling_module}
\end{equation}
For a single MSRB, since the five DSConvs are independent, the total number of parameters $P_{\mathrm{MSRB}}$ is given as:
\begin{equation}
P_{\mathrm{MSRB}}
=
5 \times P_{\mathrm{DSConv}}
=
5 \times 4C
=
20C.
\label{eq:msrb_parameter}
\end{equation}

Based on the MSRB, an ultra-lightweight network named UltraFast-LiNET is constructed for real-time low-light image enhancement. Suppose the network contains $L$ processing stages, where $L=3$ in this work, and let $l$ denote the index of different layers, $l=1,2,\ldots,L$. The encoder is defined as a stack of $L$ $M^{\downarrow}(X)$ blocks, each of which halves the resolution of its input feature map and transforms its representation. The output $X^{(l)}$ of the $l$-th encoder layer is formulated as:
\begin{equation}
\begin{aligned}
X^{(l)}
=
M^{(l)\downarrow}
\left(
X^{(l-1)}
\right),
\quad
\mathrm{for}\ l=1,2,\ldots,L,
\end{aligned}
\label{eq:encoder_layer}
\end{equation}
where $X^{(0)}=X$ is the original input to the network.

\begin{table*}[width=\textwidth,cols=8,pos=t]
\caption{Performance comparison on the LOL dataset. Dark, medium, and light blue indicate the first, second, and third best results for each metric, respectively.}
\label{tbl1}
\scriptsize
\setlength{\tabcolsep}{3pt}
\begin{tabular*}{\tblwidth}{@{} LCCCCCCC@{} }
\toprule
Method & SSIM$\uparrow$ & PSNR$\uparrow$ & LPIPS$\downarrow$ & NIQE$\downarrow$ & LOE$\downarrow$ & DE$\uparrow$ & EME$\uparrow$ \\
\midrule
ZeroDCE++~\citep{Li2021ZeroDCE} & 0.61 & 15.38 & 0.32 & 5.86 & 25.92 & 1.99 & 19.57 \\
ZeroDCE~\citep{Guo2020ZeroDCE} & 0.60 & 14.98 & 0.31 & 5.81 & 23.79 & 1.84 & 19.54 \\
SCI~\citep{Ma2022SCI} & 0.56 & 13.87 & 0.34 & 6.01 & \rankthreecell{6.81} & 1.92 & \rankthreecell{20.19} \\
RUAS~\citep{Liu2021RetinexUnrolling} & 0.46 & 11.33 & 0.36 & 5.77 & \ranktwocell{0.33} & 1.47 & \ranktwocell{20.84} \\
EnlightenGAN~\citep{Jiang2021EnlightenGAN} & 0.72 & 17.04 & 0.30 & 4.21 & 64.76 & 1.65 & 2.99 \\
FMR-NET~\citep{Chen2024FMR} & \rankonecell{0.79} & 18.87 & 0.25 & 4.63 & 27.37 & \rankthreecell{2.28} & 2.97 \\
LIME~\citep{Guo2016LIME} & 0.50 & 14.01 & 0.41 & 5.63 & 102.15 & 1.84 & \rankonecell{22.26} \\
FRR-NET~\citep{Chen2024FRR} & 0.75 & 18.27 & 0.28 & 5.78 & 24.22 & 1.94 & 3.82 \\
UTV-NET~\citep{Zheng2021AdaptiveUnfolding} & 0.74 & 15.62 & 0.20 & 4.32 & 30.54 & 2.03 & 4.57 \\
ChebyLighter~\citep{Pan2022ChebyLighter} & \ranktwocell{0.78} & 19.65 & 0.19 & 4.37 & 18.21 & 2.16 & 3.58 \\
EFI-NET~\citep{Liu2022EFINet} & 0.64 & 13.48 & 0.30 & 4.43 & 25.18 & 1.64 & 5.35 \\
PairLIE~\citep{Fu2023SimpleEnhancer} & \rankthreecell{0.77} & 18.58 & 0.23 & \ranktwocell{4.08} & 51.88 & 1.91 & 3.75 \\
NoiSER~\citep{Zhang2024NoiseSelfRegression} & 0.59 & 17.36 & 0.32 & 6.68 & 38.01 & 2.11 & 6.99 \\
URetinex-NET~\citep{Wu2022URetinex} & 0.73 & \rankthreecell{19.71} & 0.35 & 4.37 & 52.19 & \ranktwocell{2.35} & 2.56 \\
ZeroIG~\citep{Shi2024ZEROIG} & 0.50 & 17.98 & \rankthreecell{0.17} & \rankonecell{4.05} & 29.98 & 2.12 & 3.45 \\
LiteIE~\citep{Bai2025LiteIE} & 0.61 & \ranktwocell{19.73} & \rankonecell{0.11} & 4.29 & \rankonecell{0.29} & 1.98 & 20.09 \\
UltraFast-LiNETMax & 0.73 & \rankonecell{19.81} & \ranktwocell{0.14} & \rankthreecell{4.11} & 11.23 & \rankonecell{2.54} & 19.57 \\
\bottomrule
\end{tabular*}
\end{table*}

The bottleneck layer consists of only a single MSRB, and its output $X^{(L+1)}$ can be defined as follows:
\begin{equation}
X^{(L+1)}
=
\mathrm{MSRB}
\left(
X^{(L)}
\right).
\label{eq:bottleneck}
\end{equation}
The decoder progressively restores the resolution through upsampling blocks and skip connections. The output $O^{(m)}$ of the $m$-th decoder layer, where $m=L,L-1,\ldots,1$, is the sum of the upsampled features and the skip connection features from the $(m-1)$-th encoder layer, and it is formulated as:
\begin{equation}
O^{(m)}
=
M^{\uparrow}
\left(
O^{(m-1)}, X^{(m-1)}
\right)
\oplus
X^{(m-1)}.
\label{eq:decoder_layer}
\end{equation}
Finally, UltraFast-LiNET produces $L$ outputs for subsequent loss function training supervision. When $\kappa=5$, UltraFast-LiNET contains only 180 learnable parameters, with a total computation of 14.036M, denoted as UltraFast-LiNETMax hereafter. When $\kappa=1$, UltraFast-LiNET contains only 36 learnable parameters, with a total computation of 2.807M, denoted as UltraFast-LiNETMini hereafter.
\subsection{Loss Function}
\label{subsec:loss_function}

\begin{figure*}[!t]
  \centering
  \includegraphics[width=\linewidth]{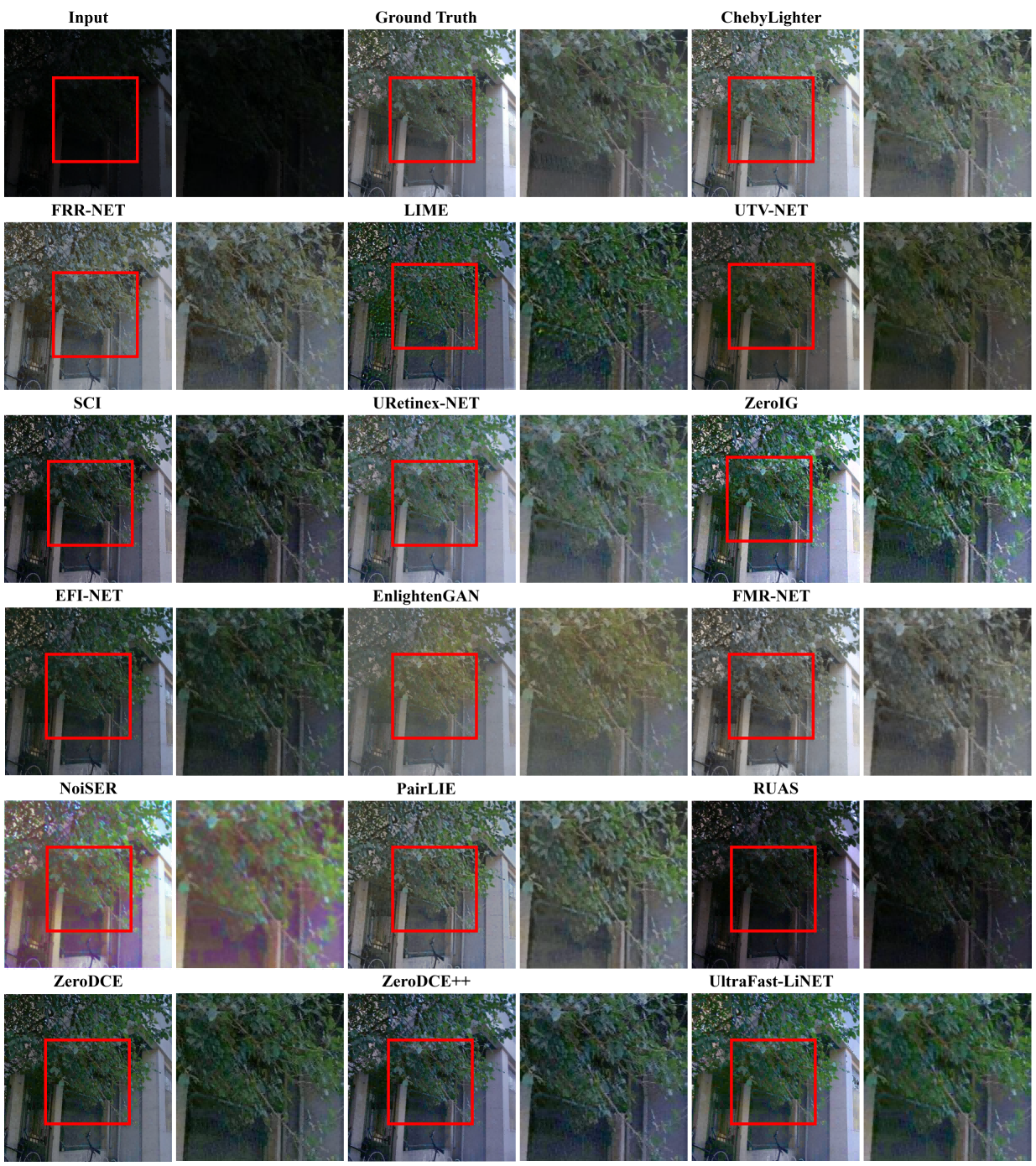}
  \caption{Visual comparison of UltraFast-LiNET with state-of-the-art methods on the LOL dataset, showing detail restoration quality with red box crop enlargements.}
  \label{fig8}
\end{figure*}

\begin{table*}[width=\textwidth,cols=8,pos=!t]
    \caption{Performance comparison on the LSRW-HUAWEI dataset. Dark, medium, and light blue indicate the first, second, and third best results for each metric, respectively.}
    \label{tbl2}
    \scriptsize
    \setlength{\tabcolsep}{3pt}
    \begin{tabular*}{\tblwidth}{@{} LCCCCCCC@{} }
    \toprule
    Method & SSIM$\uparrow$ & PSNR$\uparrow$ & LPIPS$\downarrow$ & NIQE$\downarrow$ & LOE$\downarrow$ & DE$\uparrow$ & EME$\uparrow$ \\
    \midrule
    ZeroDCE++~\citep{Li2021ZeroDCE} & 0.56 & 15.74 & 0.38 & 3.50 & 9.94 & 2.40 & 37.17 \\
    ZeroDCE~\citep{Guo2020ZeroDCE} & 0.55 & 15.57 & 0.37 & 3.81 & 8.36 & 2.27 & 37.71 \\
    SCI~\citep{Ma2022SCI} & 0.50 & 14.12 & 0.44 & 4.12 & \rankthreecell{2.19} & 1.90 & \rankthreecell{38.63} \\
    RUAS~\citep{Liu2021RetinexUnrolling} & 0.38 & 11.49 & 0.54 & 5.04 & \ranktwocell{1.90} & 1.17 & \rankonecell{43.70} \\
    EnlightenGAN~\citep{Jiang2021EnlightenGAN} & 0.63 & 18.28 & 0.42 & 3.75 & 62.62 & 1.92 & 2.17 \\
    FMR-NET~\citep{Chen2024FMR} & \ranktwocell{0.69} & \ranktwocell{20.35} & 0.39 & 3.97 & 41.25 & 3.01 & 4.49 \\
    LIME~\citep{Guo2016LIME} & 0.40 & 14.18 & 0.47 & 4.89 & 130.84 & 2.76 & \ranktwocell{42.71} \\
    FRR-NET~\citep{Chen2024FRR} & 0.66 & 19.90 & \rankonecell{0.30} & 3.77 & 3.21 & 2.76 & 7.40 \\
    UTV-NET~\citep{Zheng2021AdaptiveUnfolding} & 0.66 & 18.27 & \rankthreecell{0.35} & \ranktwocell{3.05} & 15.84 & 2.67 & 9.52 \\
    ChebyLighter~\citep{Pan2022ChebyLighter} & \rankthreecell{0.67} & 19.63 & 0.39 & \rankthreecell{3.06} & 46.86 & 2.98 & 7.08 \\
    EFI-NET~\citep{Liu2022EFINet} & 0.56 & 13.99 & 0.43 & 3.49 & 16.47 & 1.91 & 7.99 \\
    PairLIE~\citep{Fu2023SimpleEnhancer} & 0.66 & 19.07 & 0.42 & 4.32 & 42.15 & 2.51 & 5.40 \\
    NoiSER~\citep{Zhang2024NoiseSelfRegression} & 0.63 & 17.43 & 0.61 & 5.38 & 85.02 & 1.39 & 1.34 \\
    URetinex-NET~\citep{Wu2022URetinex} & \rankthreecell{0.67} & \rankonecell{20.51} & \rankthreecell{0.35} & \rankonecell{2.85} & 19.47 & \ranktwocell{3.20} & 11.00 \\
    ZeroIG~\citep{Shi2024ZEROIG} & 0.52 & 17.72 & \rankthreecell{0.35} & 4.65 & 5.70 & 2.42 & 38.02 \\
    LiteIE~\citep{Bai2025LiteIE} & 0.47 & 16.82 & 0.37 & 3.71 & \rankonecell{1.85} & \rankthreecell{3.08} & 13.54 \\
    UltraFast-LiNETMax & \rankonecell{0.71} & \rankthreecell{19.92} & \ranktwocell{0.32} & 3.52 & 9.54 & \rankonecell{3.70} & 16.05 \\
    \bottomrule
    \end{tabular*}
   
    \vspace{2em} 

    \caption{Performance comparison on the LSRW-NIKON dataset. Dark, medium, and light blue indicate the first, second, and third best results for each metric, respectively.}
    \label{tbl3}
    \scriptsize
    \setlength{\tabcolsep}{3pt}
    \begin{tabular*}{\tblwidth}{@{} LCCCCCCC@{} }
    \toprule
    Method & SSIM$\uparrow$ & PSNR$\uparrow$ & LPIPS$\downarrow$ & NIQE$\downarrow$ & LOE$\downarrow$ & DE$\uparrow$ & EME$\uparrow$ \\
    \midrule
    ZeroDCE++~\citep{Li2021ZeroDCE} & \rankonecell{0.53} & 16.71 & \rankthreecell{0.28} & 2.62 & 29.26 & 1.74 & 8.79 \\
    ZeroDCE~\citep{Guo2020ZeroDCE} & 0.42 & 16.04 & 0.29 & 2.85 & 70.83 & 1.52 & 8.87 \\
    SCI~\citep{Ma2022SCI} & 0.37 & 16.15 & 0.32 & 2.73 & \ranktwocell{15.01} & 1.76 & \rankthreecell{10.76} \\
    RUAS~\citep{Liu2021RetinexUnrolling} & 0.34 & 13.51 & 0.35 & 2.62 & \rankonecell{0.57} & 1.45 & \rankonecell{11.65} \\
    EnlightenGAN~\citep{Jiang2021EnlightenGAN} & 0.43 & 15.63 & 0.38 & 3.09 & 97.70 & 0.33 & 1.96 \\
    FMR-NET~\citep{Chen2024FMR} & \ranktwocell{0.51} & \rankonecell{18.19} & 0.31 & 2.99 & 32.85 & \rankthreecell{1.77} & 4.54 \\
    LIME~\citep{Guo2016LIME} & 0.29 & 14.76 & 0.35 & 3.38 & 97.35 & 1.63 & \ranktwocell{11.10} \\
    FRR-NET~\citep{Chen2024FRR} & 0.48 & 17.07 & 0.30 & 2.66 & \rankthreecell{15.18} & 1.54 & 5.55 \\
    UTV-NET~\citep{Zheng2021AdaptiveUnfolding} & 0.47 & 16.13 & 0.32 & 3.22 & 19.21 & 1.22 & 6.62 \\
    ChebyLighter~\citep{Pan2022ChebyLighter} & 0.42 & 15.38 & 0.30 & 2.73 & 37.50 & 1.65 & 4.49 \\
    EFI-NET~\citep{Liu2022EFINet} & 0.44 & 15.51 & 0.29 & \rankthreecell{2.42} & 46.72 & 1.44 & 7.56 \\
    PairLIE~\citep{Fu2023SimpleEnhancer} & 0.45 & 17.18 & \ranktwocell{0.27} & 3.68 & 75.00 & 1.59 & 5.53 \\
    NoiSER~\citep{Zhang2024NoiseSelfRegression} & 0.45 & 15.92 & 0.43 & 3.77 & 78.71 & 1.55 & 3.27 \\
    URetinex-NET~\citep{Wu2022URetinex} & 0.47 & \rankthreecell{17.73} & \rankonecell{0.26} & \rankonecell{2.32} & 35.28 & 1.62 & 4.60 \\
    ZeroIG~\citep{Shi2024ZEROIG} & 0.30 & 14.51 & 0.34 & 2.83 & 19.93 & \rankonecell{2.12} & 9.42 \\
    LiteIE~\citep{Bai2025LiteIE} & 0.41 & 16.02 & 0.41 & 3.61 & 33.60 & 1.52 & 8.23 \\
    UltraFast-LiNETMax & \rankthreecell{0.50} & \ranktwocell{17.76} & 0.32 & \ranktwocell{2.39} & 31.04 & \ranktwocell{1.79} & 9.44 \\
    \bottomrule
    \end{tabular*}
\end{table*}

\begin{table*}[width=\textwidth,cols=8,pos=!t]
    \caption{Performance comparison on the LoLI-Street dataset. Dark, medium, and light blue indicate the first, second, and third best results for each metric, respectively.}
    \label{tbl4}
    \scriptsize
    \setlength{\tabcolsep}{3pt}
    \begin{tabular*}{\tblwidth}{@{} LCCCCCCC@{} }
    \toprule
    Method & SSIM$\uparrow$ & PSNR$\uparrow$ & LPIPS$\downarrow$ & NIQE$\downarrow$ & LOE$\downarrow$ & DE$\uparrow$ & EME$\uparrow$ \\
    \midrule
    ZeroDCE++~\citep{Li2021ZeroDCE} & 0.87 & 13.87 & 0.23 & 3.48 & 60.09 & 0.84 & 4.09 \\
    ZeroDCE~\citep{Guo2020ZeroDCE} & 0.84 & 13.31 & 0.22 & 3.29 & 68.16 & 0.23 & 3.94 \\
    SCI~\citep{Ma2022SCI} & 0.86 & 14.93 & 0.21 & \rankonecell{3.09} & 24.03 & 1.06 & \rankthreecell{5.26} \\
    RUAS~\citep{Liu2021RetinexUnrolling} & 0.81 & 12.43 & 0.24 & 3.21 & \rankonecell{0.05} & 1.36 & \rankonecell{6.26} \\
    EnlightenGAN~\citep{Jiang2021EnlightenGAN} & \rankthreecell{0.90} & 17.18 & 0.15 & 3.17 & 98.38 & 1.21 & 2.71 \\
    FMR-NET~\citep{Chen2024FMR} & 0.85 & 13.77 & 0.23 & 3.94 & \ranktwocell{5.10} & 0.92 & 2.12 \\
    LIME~\citep{Guo2016LIME} & 0.74 & 11.54 & 0.36 & 3.31 & 151.06 & 0.98 & \ranktwocell{5.35} \\
    FRR-NET~\citep{Chen2024FRR} & 0.87 & 14.77 & 0.20 & 3.58 & 7.56 & 0.72 & 2.32 \\
    UTV-NET~\citep{Zheng2021AdaptiveUnfolding} & 0.74 & 9.54 & 0.28 & 3.75 & 26.16 & 0.74 & 3.39 \\
    ChebyLighter~\citep{Pan2022ChebyLighter} & \ranktwocell{0.91} & \ranktwocell{20.91} & \rankonecell{0.12} & \ranktwocell{3.11} & \rankthreecell{6.61} & 1.19 & 2.02 \\
    EFI-NET~\citep{Liu2022EFINet} & 0.86 & 13.72 & 0.18 & \rankthreecell{3.14} & 27.36 & 1.09 & 3.07 \\
    PairLIE~\citep{Fu2023SimpleEnhancer} & 0.88 & 18.30 & 0.19 & 3.45 & 105.06 & 0.72 & 2.26 \\
    NoiSER~\citep{Zhang2024NoiseSelfRegression} & 0.77 & 14.32 & 0.44 & 4.41 & 53.22 & \rankthreecell{1.38} & 2.53 \\
    URetinex-NET~\citep{Wu2022URetinex} & 0.89 & 18.56 & \rankthreecell{0.14} & 3.97 & 26.02 & 0.69 & 2.29 \\
    ZeroIG~\citep{Shi2024ZEROIG} & 0.82 & 17.50 & 0.22 & 3.19 & 36.00 & 1.32 & 4.40 \\
    LiteIE~\citep{Bai2025LiteIE} & 0.89 & \rankthreecell{19.72} & 0.18 & 3.40 & 11.87 & \ranktwocell{1.41} & 3.09 \\
    UltraFast-LiNETMax & \rankonecell{0.92} & \rankonecell{25.51} & \ranktwocell{0.13} & 3.42 & 30.51 & \rankonecell{1.58} & 2.73 \\
    \bottomrule
    \end{tabular*}
    
    \vspace{2em}

    \caption{Computational efficiency comparison on the LOL dataset. Dark, medium, and light blue indicate the first, second, and third best results for each metric, respectively.}
    \label{tbl5}
    \scriptsize
    \setlength{\tabcolsep}{3pt}
    \begin{tabular*}{\tblwidth}{@{} LCCCCC@{} }
    \toprule
    Method & SSIM$\uparrow$ & PSNR$\uparrow$ & Params (K)$\downarrow$ & FLOPs (G)$\downarrow$ & Runtimes (ms)$\downarrow$ \\
    \midrule
    ZeroDCE++~\citep{Li2021ZeroDCE} & 0.61 & 15.38 & 10.6 & 0.33 & 3.99 \\
    ZeroDCE~\citep{Guo2020ZeroDCE} & 0.60 & 14.98 & 79.4 & 5.21 & 7.03 \\
SCI~\citep{Ma2022SCI} & 0.56 & 13.87 & 0.3 & 0.0619 & 1.98 \\
    RUAS~\citep{Liu2021RetinexUnrolling} & 0.46 & 11.33 & 1.4 & 0.2813 & 10.1 \\
    EnlightenGAN~\citep{Jiang2021EnlightenGAN} & 0.72 & 17.04 & 8636 & 61.01 & 10.6 \\
    FMR-NET~\citep{Chen2024FMR} & \rankonecell{0.79} & 18.87 & 196.8 & 102.8 & 116 \\
    LIME~\citep{Guo2016LIME} & 0.50 & 14.01 & N/A & N/A & 758 \\
    FRR-NET~\citep{Chen2024FRR} & 0.75 & 18.27 & 12.21 & 0.216 & 75.7 \\
    UTV-NET~\citep{Zheng2021AdaptiveUnfolding} & 0.74 & 15.62 & 7745 & 58.29 & 46.5 \\
    ChebyLighter~\citep{Pan2022ChebyLighter} & \ranktwocell{0.78} & 19.65 & 73 & 17.25 & 76.4 \\
    EFI-NET~\citep{Liu2022EFINet} & 0.64 & 13.48 & 129.2 & 9.38 & 44.8 \\
    PairLIE~\citep{Fu2023SimpleEnhancer} & \rankthreecell{0.77} & 18.58 & 34.18 & 22.35 & 28.4 \\
    NoiSER~\citep{Zhang2024NoiseSelfRegression} & 0.59 & 17.36 & 1.763 & 8.62 & 3.02 \\
    URetinex-NET~\citep{Wu2022URetinex} & 0.73 & \rankthreecell{19.71} & 838.3 & 136.01 & 36.9 \\
    ZeroIG~\citep{Shi2024ZEROIG} & 0.50 & 17.98 & 123.63 & 118.73 & 17.2 \\
    LiteIE~\citep{Bai2025LiteIE} & 0.61 & \ranktwocell{19.73} & \ranktwocell{0.058} & \rankthreecell{0.00389} & \ranktwocell{1.83} \\
    UltraFast-LiNETMini & 0.51 & 17.19 & \rankonecell{0.036} & \rankonecell{3.0$\times 10^{-6}$} & \rankonecell{1.72} \\
UltraFast-LiNETMax & 0.73 & \rankonecell{19.81} & \rankthreecell{0.18} & \ranktwocell{1.4$\times 10^{-5}$} & \rankthreecell{1.89} \\
    \bottomrule
    \end{tabular*}
\end{table*}

To effectively train UltraFast-LiNET, a composite loss function $\ell_{\mathrm{total}}$ is designed to optimize the quality of the enhanced images from multiple perspectives. This loss function consists of three components: a reconstruction loss $\ell_{\mathrm{rec}}$, a structural similarity loss $\ell_{\mathrm{ms\text{-}ssim}}$, and a novel multi-level gradient perception loss $\ell_{\mathrm{grad}}$. The total loss function $\ell_{\mathrm{total}}$ is defined as the weighted sum of these three components:
\begin{equation}
\ell_{\mathrm{total}}
=
\lambda_{\mathrm{rec}}\ell_{\mathrm{rec}}
+
\lambda_{\mathrm{ms\text{-}ssim}}\ell_{\mathrm{ms\text{-}ssim}}
+
\lambda_{\mathrm{grad}}\ell_{\mathrm{grad}},
\label{eq:total_loss}
\end{equation}

where $\lambda_{\mathrm{rec}}$, $\lambda_{\mathrm{ms\text{-}ssim}}$, and $\lambda_{\mathrm{grad}}$ represent the hyperparameters of each loss. Referring to the design concept of $L_1$ and SSIM hybrid loss, $\lambda_{\mathrm{rec}}$ and $\lambda_{\mathrm{ms\text{-}ssim}}$ are 0.975 and 0.025, respectively \citep{Zhao2015LossFunctions}. The empirical value of $\lambda_{\mathrm{grad}}$ is 1; see the ablation experiment of the loss function.
The objective of the reconstruction loss is to minimize the pixel-level difference between the enhanced image and the normal-light image. A smooth $L_1$ loss is adopted as the reconstruction loss function $\ell_{\mathrm{rec}}$. It combines the advantages of $L_1$ and $L_2$ losses, is insensitive to outliers, and provides stable gradients when the error is small. Let the enhanced image be denoted as $E$ and the normal-light image as $G$. The reconstruction loss $\ell_{\mathrm{rec}}$ is defined as:
\begin{equation}
\ell_{\mathrm{rec}}(E,G)
=
\frac{1}{N}
\sum_{i=1}^{N}
\mathrm{smooth}_{L1}(e_i-g_i),
\label{eq:rec_loss}
\end{equation}

where $e_i$ and $g_i$ represent the $i$-th pixel value of image $E$ and image $G$, respectively. $N$ is the total pixel value of the image. The $\mathrm{smooth}_{L1}$ function can be defined as follows:
\begin{equation}
\mathrm{smooth}_{L1}(x)
=
\begin{cases}
0.5x^2, & \mathrm{if}\ |x|<1, \\
|x|-0.5, & \mathrm{otherwise}.
\end{cases}
\label{eq:smooth_l1}
\end{equation}

Pixel-level loss functions such as $L_1$ or $L_2$ are inadequate for capturing structural information consistent with human visual perception. To address this shortcoming, the multi-scale structural similarity (MS-SSIM) index is incorporated as a loss component. This metric assesses perceptual similarity in luminance, contrast, and structure across multiple scales. The MS-SSIM loss is defined as the difference between 1 and the MS-SSIM value, forming a minimization objective expressed as:
\begin{equation}
\ell_{\mathrm{ms\text{-}ssim}}(E,G)
=
1-\mathrm{MS\text{-}SSIM}(E,G).
\label{eq:ms_ssim_loss}
\end{equation}

The UltraFast-LiNET framework is extremely light\-weight, leading to feature degradation during propagation. To mitigate this issue, a multi-level gradient-aware loss function $\ell_{\mathrm{grad}}$ is proposed. This loss enforces gradient consistency constraints across multiple decoder stages, effectively preserving structural information from coarse to fine granularity.

Assume that $\{O_m\}_{m=1}^{3}$ is the feature map output by the network decoder at three different scales, where $O_1$ is the final enhanced image $E$. Therefore, $\ell_{\mathrm{grad}}$ can be expressed as:
\begin{equation}
\resizebox{0.88\columnwidth}{!}{$
\begin{aligned}
\ell_{\mathrm{grad}}\!\left(\{O_m\}_{m=1}^{3},G\right)
&= \sum_{k=1}^{3} \omega_k \Big[
\ell_{\mathrm{rec}}\!\left(\nabla_x(O_m), \nabla_x\left(D_k(G)\right)\right)\\
&\quad + \ell_{\mathrm{rec}}\!\left(\nabla_y(O_m), \nabla_y\left(D_k(G)\right)\right) \Big],
\end{aligned}
$}
\label{eq:grad_loss}
\end{equation}

where $\omega_k$ is a hyperparameter that controls the weight of the loss at different scales. Based on the model architecture and experimental validation, $\omega_1=1.0$, $\omega_2=1.0$, and $\omega_3=0.04$ are assigned; refer to the ablation analysis. $D_k$ denotes a downsampling operator that applies bilinear interpolation to adjust the spatial resolution of the input image $G$ to exactly match that of the intermediate feature map $O_m$. This ensures that gradient comparison is carried out at the same dimensional level. $\nabla_x$ and $\nabla_y$ represent the gradient calculation operations in the horizontal and vertical directions, respectively. We use the standard $3 \times 3$ Sobel operator to implement it, which is equivalent to convolution with the following convolution kernels $K_x$ and $K_y$:
\begin{equation}
K_x
=
\begin{bmatrix}
-1 & 0 & 1 \\
-2 & 0 & 2 \\
-1 & 0 & 1
\end{bmatrix},
\qquad
K_y
=
\begin{bmatrix}
-1 & -2 & -1 \\
0 & 0 & 0 \\
1 & 2 & 1
\end{bmatrix}.
\label{eq:sobel_kernels}
\end{equation}
Before applying the operator, all images and feature maps are converted into single-channel grayscale representations so that the loss function focuses on the preservation of luminance gradients.
\subsection{Implementation details}
\label{subsec:implementation_details}

The overall architecture of UltraFast-LiNET is implemented using PyTorch and the network is trained using an NVIDIA GeForce RTX 4060 Laptop GPU and a 12th Gen Intel Core i7-12650H processor at 2.30 GHz \citep{Paszke2019PyTorch}. Training is conducted for 360 epochs with the Adam optimizer, an initial learning rate of 0.01, and a batch size of 40. The learning rate is multiplied by a decay factor $\Upsilon=0.1$ every 40 epochs \citep{Kingma2015Adam}. For the LOL and LSRW datasets, we strictly follow their official train/test splits for training and evaluation. For the LoLI-Street dataset, which provides no official partition, a 9:1 random train/test split is applied. We center-crop both low-light and normal-light images into $180 \times 180$ pixel patches to maintain consistent input dimensions. We subsequently employ the ToTensor transformation to normalize pixel values from the integer range into the $[0.0, 1.0]$ floating-point range and convert the data structure into the PyTorch Tensor format.

\begin{figure*}[t]
  \centering
  \includegraphics[width=\linewidth]{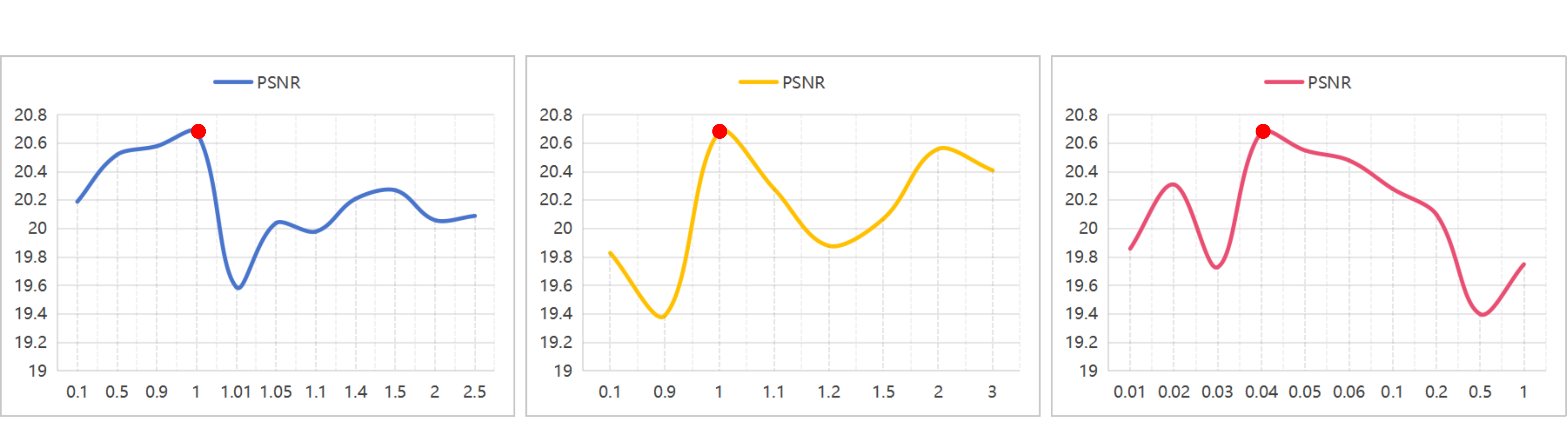}
  \caption{Ablation study on the hyperparameter configuration of the multi-level gradient-aware loss function $\ell_{\mathrm{grad}}$ conducted on the LOL dataset. The optimal weights $\omega_k$ maximizing PSNR are identified.}
  \label{fig9}
\end{figure*}

\section{Experiments}
\label{sec:experiments_and_results}

Our experiments select the LOL, LSRW-HUAWEI, LSRW-NIKON, and LoLI-Street datasets for the training and testing of UltraFast-LiNET \citep{Wei2018RetinexNet,Hai2023R2RNet,Islam2024LOLIStreet}. The LOL dataset is the first paired dataset for supervised learning in low-light image enhancement, and it mainly consists of both synthetic data and real captured data. The LSRW dataset is the first large-scale real-world paired low-light/normal-light image dataset, containing two different paired subsets captured with HUAWEI P40 Pro and NIKON D7500 devices. Finally, LoLI-Street is the first large-scale low-light image dataset for driving scenarios.

\begin{table}[width=\linewidth,cols=5,pos=t]
\caption{Ablation study on loss function selection conducted on the LoLI-Street dataset. Dark, medium, and light blue indicate the first, second, and third best results, respectively.}
\label{tbl6}
\scriptsize
\setlength{\tabcolsep}{3pt}
\begin{tabular*}{\tblwidth}{@{} LCCCC@{} }
\toprule
Smooth L1 & MS-SSIM & Grad & PSNR$\uparrow$ & SSIM$\uparrow$ \\
\midrule
$\checkmark$ & $\checkmark$ & $\checkmark$ & \rankonecell{24.98} & \rankonecell{0.86} \\
$\checkmark$ & $\checkmark$ & $\times$ & \ranktwocell{19.71} & 0.75 \\
$\times$ & $\times$ & $\checkmark$ & 17.53 & \rankthreecell{0.77} \\
$\checkmark$ & $\times$ & $\times$ & 18.46 & 0.69 \\
$\checkmark$ & $\times$ & $\checkmark$ & \rankthreecell{18.91} & \ranktwocell{0.82} \\
$\times$ & $\checkmark$ & $\checkmark$ & 17.89 & 0.71 \\
$\times$ & $\checkmark$ & $\times$ & 17.38 & 0.71 \\
\bottomrule
\end{tabular*}
\end{table}

\subsection{Comparative LLIE Results}
\label{subsec:comparative_llie_results}

We select state-of-the-art methods in the LLIE field for comparison with UltraFast-LiNET. Their evaluation results on the LOL, LSRW-HUAWEI, LSRW-NIKON, and LoLI-Street datasets are reported in Tables~\ref{tbl1}--\ref{tbl4} \citep{Islam2024LOLIStreet,Guo2016LIME,Li2021ZeroDCE,Guo2020ZeroDCE,Ma2022SCI,Liu2021RetinexUnrolling,Jiang2021EnlightenGAN,Chen2024FMR,Chen2024FRR,Zheng2021AdaptiveUnfolding,Pan2022ChebyLighter,Liu2022EFINet,Fu2023SimpleEnhancer,Zhang2024NoiseSelfRegression,Wu2022URetinex,Shi2024ZEROIG,Wei2018RetinexNet,Hai2023R2RNet}. The evaluation metrics mainly include full-reference image metrics PSNR, SSIM, and LPIPS, as well as no-reference image metrics NIQE, LOE, DE, and EME. Except for Table~\ref{tbl2}, our method achieves top-two PSNR and SSIM results on all four datasets. Similarly, the proposed method also obtains excellent rankings in the remaining metrics.

To further verify the computational effectiveness of UltraFast-LiNET, both the proposed method and state-of-the-art approaches are deployed on a Jetson AGX Orin 64GB platform. The device features an NVIDIA Ampere architecture GPU with 2048 CUDA cores and 64 Tensor Cores, accompanied by a 12-core Arm Cortex-A78AE v8.2 64-bit CPU. The parameter counts and computational efficiency of UltraFast-LiNET and comparative methods are presented in Table~\ref{tbl5}. We select the LOL dataset validation set as the benchmark, obtaining average inference time per image for each method. Our approach achieves leading performance across all evaluation metrics. These results are remarkable considering that the model size of UltraFast-LiNET is significantly smaller than current state-of-the-art methods.

\begin{figure}[!t]
  \centering
  \includegraphics[width=\linewidth]{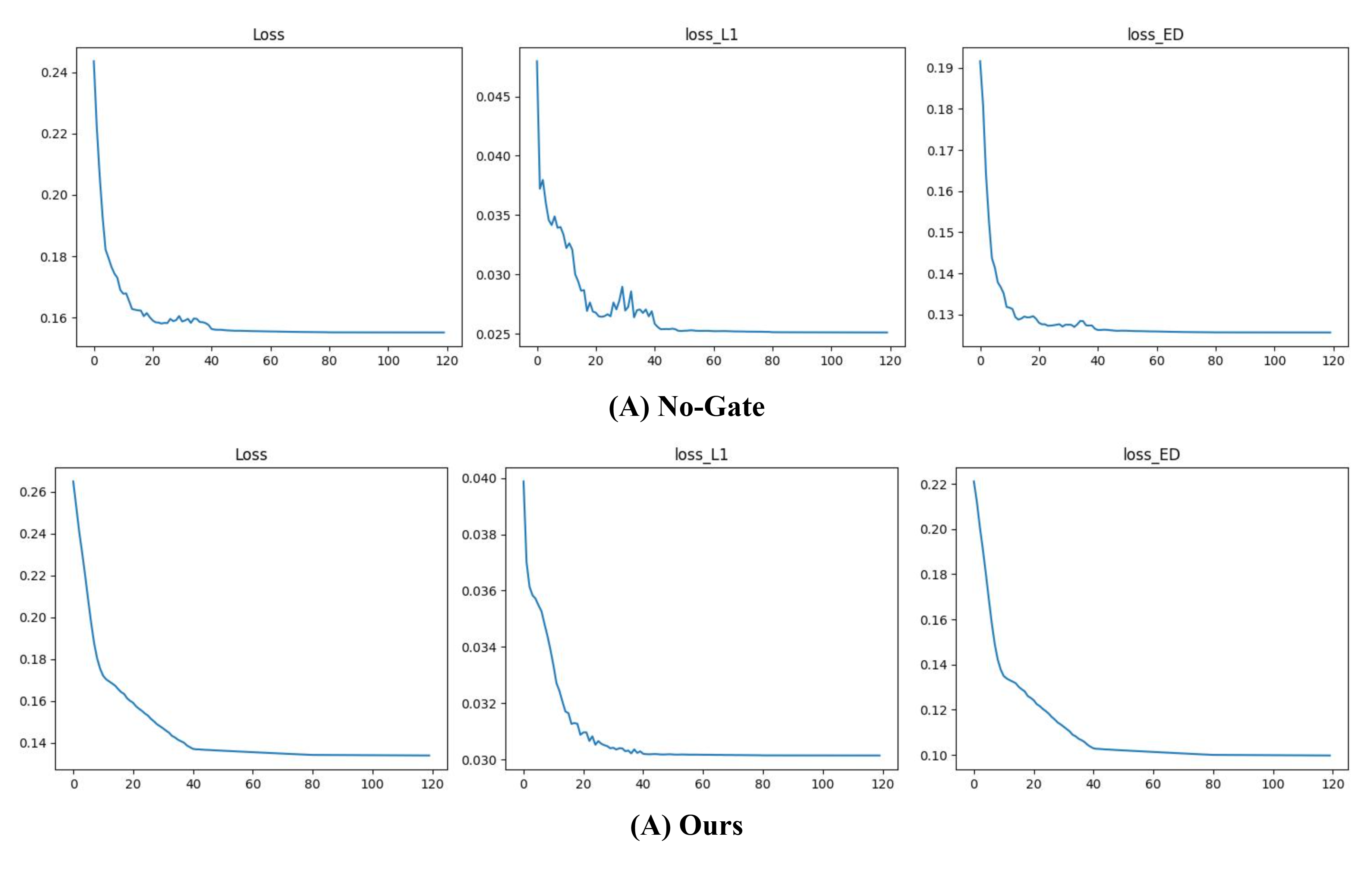}
  \caption{Training curves for the DSConv Sigmoid gating ablation study on the LOL-v1 dataset. The gated variant converges more stably and achieves higher validation PSNR, demonstrating the key role of dynamic gating in adaptive information filtering.}
  \label{fig10}
\end{figure}

Figs.~\ref{fig4}--\ref{fig7} shows visual comparison results of images selected from the LOL, LSRW-HUAWEI, LSRW-NIKON, and LoLI-Street datasets. From all images, the success of our method in color restoration can be observed. In the first image, UltraFast-LiNET is almost the most successful in recovering brightness and color. Although the saturation of our result is lower than that of ZeroIG, it is closer to the reference image. The second image verifies that UltraFast-LiNET is more balanced in global image processing and does not suffer from uneven brightness similar to EnlightenGAN. Although PairLIE and ZeroIG produce higher saturation and stronger denoising ability, our method is closer to the reference image, which is verified in the third image. The image enhanced by ZeroIG is overly bright, leading to severe loss of cloud and fog details, whereas our method effectively preserves these details. The fourth image verifies that under extremely low brightness, our method does not produce image color block stratification.

As shown in Fig.~\ref{fig8}, to further verify the advantage of UltraFast-LiNET in image detail restoration, we select more complex scenes from the LOL dataset for detailed visual verification. It is evident that although the saturation of UltraFast-LiNET is higher than that of the reference image, its detail restoration remains at a leading level. Problems such as insufficient brightness, color distortion, and over-sharpening do not appear.

In conclusion, while UltraFast-LiNET demonstrates some generalization limitations in enhancing certain images, its performance is notable given a maximum of 180 learnable parameters. The method effectively balances image quality with computational efficiency and achieves superior performance compared to state-of-the-art approaches.
\begin{table}[width=\linewidth,cols=3,pos=t]
\caption{Ablation study on the impact of different numbers and dia settings of DSConv on the model using the LoLI-Street dataset. Dark, medium, and light blue indicate the first, second, and third best results, respectively.}
\label{tbl7}
\scriptsize
\setlength{\tabcolsep}{3pt}
\begin{tabular*}{\tblwidth}{@{} LCC@{} }
\toprule
Dia & PSNR$\uparrow$ & SSIM$\uparrow$ \\
\midrule
0 & 21.45 & 0.71 \\
0+1 & 22.43 & 0.74 \\
0+1+2 & 22.73 & 0.74 \\
0+1+2+3 & 23.41 & 0.81 \\
0+1+2+3+4 & \rankonecell{24.76} & \rankonecell{0.89} \\
1 & 17.80 & 0.61 \\
1+2 & 18.20 & 0.73 \\
1+2+3 & 23.63 & 0.78 \\
1+2+3+4 & \rankthreecell{23.72} & \ranktwocell{0.87} \\
2 & 16.23 & 0.59 \\
2+3 & 23.00 & 0.80 \\
2+3+4 & \ranktwocell{24.02} & \rankthreecell{0.86} \\
3 & 17.22 & 0.67 \\
3+4 & 18.09 & 0.74 \\
4 & 15.10 & 0.62 \\
\bottomrule
\end{tabular*}
\end{table}

\subsection{Ablation Study}
\label{subsec:ablation_study}

Ablation studies are conducted to evaluate the effectiveness of the loss function selection in UltraFast-LiNET, the correctness of parameter settings in the multi-level gradient-aware loss function $\ell_{\mathrm{grad}}$, and the influence of different numbers and settings of DSConv on the model. First, the effectiveness of loss function selection is investigated in Table~\ref{tbl6} using a subset of the LoLI-Street dataset. It is evident that the composite loss function achieves the best result. Meanwhile, this study also verifies that $\ell_{\mathrm{grad}}$ effectively improves the SSIM metric.

As shown in Fig.~\ref{fig9}, to quickly verify the correctness of the parameters in the multi-level gradient-aware loss function, we conduct an ablation study on the LOL dataset. Multiple rounds of parameter tuning are performed mainly on the first image in the validation set to obtain the hyperparameters that produce the best PSNR performance in UltraFast-LiNET. Referring to the previous experimental results, the correctness of hyperparameter selection makes our method perform very well.

Finally, we use images from LoLI-Street to conduct an ablation study on the influence of different numbers and settings of DSConv on the model. As shown in Table~\ref{tbl7}, PSNR and SSIM values are presented when different numbers of DSConv and different settings are used. Specifically, the notation $2+3+4$ denotes the following MSRB combination:
\begin{equation}
\resizebox{0.82\columnwidth}{!}{$\displaystyle
\mathrm{MSRB}(X)
=
\sum_{\mathrm{dia}=2}^{4}
g_{\mathrm{dia}}(X)
=
\sum_{\mathrm{dia}=2}^{4}
\mathrm{DSConv}_{\mathrm{dia}}(X).
$}
\label{eq:msrb_ablation}
\end{equation}
It is evident that the best performance is achieved when UltraFast-LiNET combines more DSConv layers. To minimize the model's parameter size, no further fusion of additional DSConv layers is performed in this work.

\begin{figure}[!t]
  \centering
  \includegraphics[width=\linewidth]{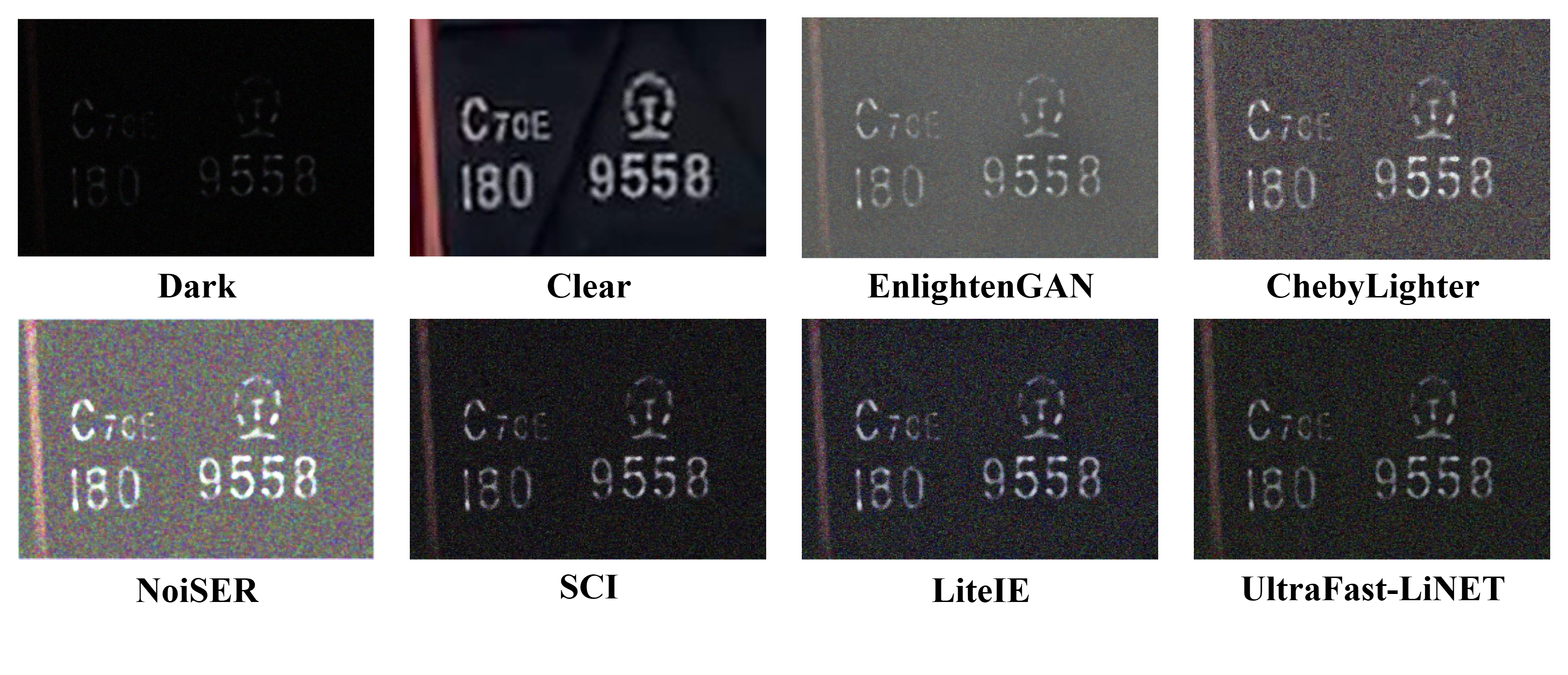}
  \caption{Qualitative comparison of different image enhancement methods on the downstream low-light train number recognition task (CRND dataset synthesized via OneRestore). Our method restores brightness while preserving clear character features.}
  \label{fig11}
\end{figure}

To validate the effectiveness of the Sigmoid gating modulation in DSConv, we conduct a No-gate ablation experiment on the LOL-v1 standard dataset using the UltraFast-LiNETMax variant. In this variant, $G \otimes X_r$ is removed and changed to $Y=X_{\mathrm{agg}}+X_r$. All other network structures, Dia combinations, and training settings remain identical. As shown in Table~\ref{tbl8}, compared with the No-gate method, the complete model improves PSNR and SSIM by 2.77~dB and 16\%, respectively. The training curves in Fig.~\ref{fig10} also show that the gated version converges more stably on the validation set and achieves higher performance, indicating that dynamic gating can effectively filter aggregated information and plays a key role in adaptive information filtering.

\subsection{Low-light Train Number Recognition}
\label{subsec:downstream_task}
\begin{figure}[!t]
  \centering
  \includegraphics[width=\linewidth]{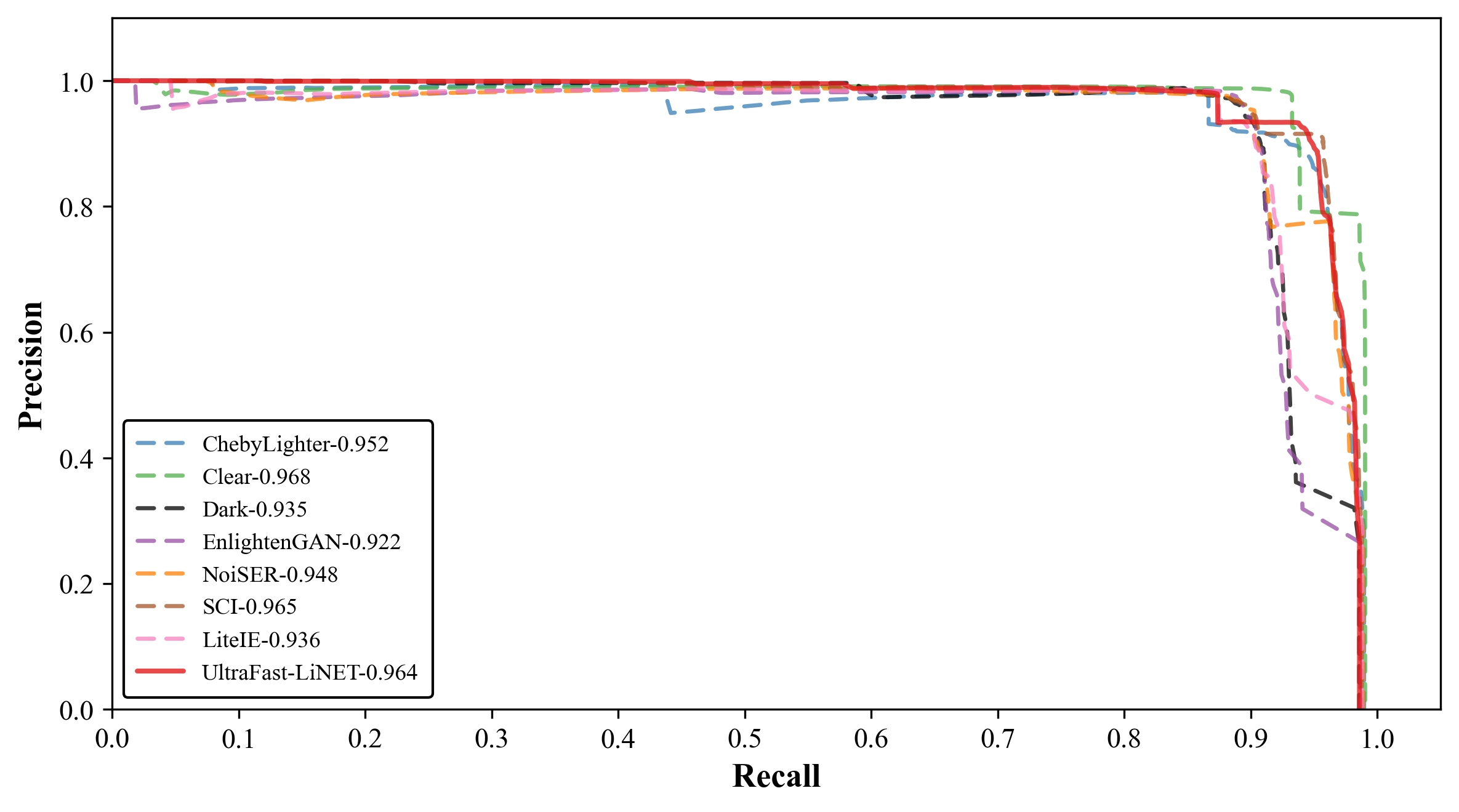}
  \caption{Precision-Recall curves of different enhancement algorithms on the downstream train number recognition task using MMGA. Legend values denote the recognition score of each method. UltraFast-LiNET (score 0.964) significantly outperforms the unprocessed dark input (0.935).}
  \label{fig12}
\end{figure}

To evaluate the indirect benefit of image enhancement on downstream high-level vision tasks, we adopt MMGA (Multi-Scale Multi-Layer Gated Attention Mechanism) as the baseline recognition algorithm. The CRND (Chinese Railway Number Dataset) is first converted into noisy low-light images using the OneRestore~\citep{Guo2024OneRestore} synthesis pipeline. Various state-of-the-art enhancement algorithms and the proposed UltraFast-LiNETMax are then applied to the synthesized data for MMGA training and testing. Fig.~\ref{fig11} presents qualitative visual comparisons of different enhancement methods. While some methods such as EnlightenGAN and ChebyLighter improve brightness, they introduce contrast degradation or blurred characters. NoiSER introduces visible color distortion and residual noise. By contrast, UltraFast-LiNET effectively restores brightness while maximally removing noise and preserving clear character features. Fig.~\ref{fig12} presents the Precision-Recall (PR) curves and quantitative results. Compared to the unprocessed synthetic input (Dark, score 0.935), our method achieves a recognition score of 0.964, demonstrating an effective improvement in downstream task performance.

\begin{table}[width=\linewidth,cols=4,pos=!t]
\caption{Ablation study on the Sigmoid gating mechanism in DSConv. The No-gate variant removes the adaptive gating branch and directly sums the shifted features with the residual.}
\label{tbl8}
\scriptsize
\setlength{\tabcolsep}{3pt}
\begin{tabular*}{\tblwidth}{@{} LCCC@{} }
\toprule
Method & Gating & PSNR$\uparrow$ & SSIM$\uparrow$ \\
\midrule
No-gate DSConv & $\times$ & 17.04 & 0.57 \\
Proposed DSConv & $\checkmark$ & \rankonecell{19.81} & \rankonecell{0.73} \\
\bottomrule
\end{tabular*}
\end{table}

\section{Conclusion}
\label{sec:conclusion}

This paper proposes UltraFast-LiNET, a lightweight multi-scale shift convolutional network for real-time low-light image enhancement. The network achieves advanced performance on multiple datasets. Targeting the limited computing power of mobile and edge embedded devices, UltraFast-LiNET provides a solution that balances image quality and computational efficiency, with only 180 learnable parameters in the full network and a minimum compressible size of 36 learnable parameters. Through extensive comparisons with SOTA algorithms, UltraFast-LiNET demonstrates its performance advantages. In future work, we will consider how to optimize the generalization ability of the algorithm so that it can adapt to more complex real-world low-light scenes.

\section*{Acknowledgements}

The authors thank the editors and reviewers for their comments, which led to the improvement of this paper. This study is supported by the National Natural Science Foundation of China (Grant No. 52272421).

\section*{Declaration of Competing Interest}
The authors declare that they have no known competing financial interests or personal relationships that could have appeared to influence the work reported in this paper.

\printcredits

\bibliographystyle{cas-model2-names-unsrt}

\bibliography{cas-refs}



\end{document}